\documentclass[12pt]{article}

\usepackage[a4paper, total={6in, 8in}]{geometry}
\usepackage[T1]{fontenc}
\usepackage{lmodern}
\usepackage{microtype}
\usepackage{graphicx}
\usepackage{subfig}
\usepackage{makecell}
\usepackage{multirow}
\usepackage{multicol}
\usepackage{svg}

\usepackage{booktabs} 
\usepackage[english]{babel}
\usepackage[utf8x]{inputenc}
\usepackage[T1]{fontenc}
\usepackage{comment}
\usepackage[export]{adjustbox}
\usepackage{booktabs}

\usepackage{amsmath}
\usepackage{graphicx}
\usepackage{amsmath}
\usepackage{bm}
\usepackage{amsthm}
\usepackage{amssymb}
\usepackage{bbm}
\usepackage{verbatim}
\usepackage{graphicx}
\usepackage{afterpage}
\usepackage{etoolbox}
\usepackage{float}
\usepackage{rotating}
\usepackage[inline]{enumitem}

\DeclareMathOperator{\erf}{erf}
\usepackage{algorithmic}
\usepackage{algorithm}
\usepackage[colorinlistoftodos]{todonotes}
\usepackage[colorlinks=true, allcolors=blue]{hyperref}

\newcommand {\myvec}[1] {{\mbox{\boldmath $#1$}}}

\newcommand{\prob}{\mathbb{P}}

\usepackage[colorinlistoftodos]{todonotes}

\newcommand{\norm}[1]{\left\lVert#1\right\rVert}
\usepackage{hyperref}

\title{Locally Sparse Neural Networks for Tabular Biomedical Data }
\date{}

\author{ Junchen Yang $^{1 \ast }$ \and Ofir Lindenbaum $^{1 \ast }$ \and Yuval Kluger$^{1 \dagger}$\\
\normalsize{$^{1}$Yale University, USA;}
\normalsize{$^\dagger$Corresponding author. E-mail: yuval.kluger@yale.edu}\\\normalsize{Address: 333 Cedar St, New Haven, CT 06510, USA}\\
\normalsize{$^\ast$ These authors contributed equally.}
}
\begin{document}
\maketitle

\begin{abstract}
 
Tabular datasets with low-sample-size or many variables are prevalent in biomedicine. Practitioners in this domain prefer linear or tree-based models over neural networks since the latter are harder to interpret and tend to overfit when applied to tabular datasets. To address these neural networks' shortcomings, we propose an intrinsically interpretable network for heterogeneous biomedical data. We design a locally sparse neural network where the local sparsity is learned to identify the subset of most relevant features for each sample. This sample-specific sparsity is predicted via a \textit{gating} network, which is trained in tandem with the \textit{prediction} network. By forcing the model to select a subset of the most informative features for each sample, we reduce model overfitting in low-sample-size data and obtain an interpretable model. We demonstrate that our method outperforms state-of-the-art models when applied to synthetic or real-world biomedical datasets using extensive experiments. Furthermore, the proposed framework dramatically outperforms existing schemes when evaluating its interpretability capabilities. Finally, we demonstrate the applicability of our model to two important biomedical tasks: survival analysis and marker gene identification.    
\end{abstract}

\section{Introduction}

Machine learning has revolutionized the way we do scientific research. In recent years, deep neural networks (NN) have closed the performance gap between humans and machines in disciplines such as vision, image processing, audio processing, and natural language processing. The tremendous success of these complex models may be explained by an increase in data size, computational resources that enable training deeper networks \cite{tishby2015deep,arora2016understanding}, or by implicit properties of the optimization tools \cite{yaguchi2018adam,soudry2018implicit}. State-of-the-art frameworks, such as convolutional neural networks, recurrent neural networks, and transformers exploit structures or invariants in the data to inform the design of the NN. Unfortunately, these models are not suitable for biomedical applications when the associated datasets are tabular, lack spatial or temporal structure, or are heterogeneous. Therefore, biomedical data pose a challenge for deep nets and require deviating from tried and true methodologies \cite{tabular,tabular2,tabular_segal,shwartz2021tabular}.


In medicine or biology, practitioners seek for ML models that are \textit{accurate} and \textit{interpretable}. Accuracy is important for improving personalized prognosis and diagnosis. At the same time, interpretability can lead to the identification of driving factors in complex high-dimensional systems and is imperative to help practitioners trust the model. When trained on tabular biomedical data, deep nets are hard to interpret and may lead to low accuracy. This is because biomedical datasets are often low-sample-size (LSS) \cite{liu2017deep,aoshima2018survey}, high dimensional, or contain nuisance features. These challenges often lead practitioners to abandon NNs and switch to linear models in biomedicine. Linear feature selection models such as \cite{Lasso,fan2001variable,lindenbaum2021randomly,lindenbaum2021refined,jana2021support} are interpretable and able to cope with high-dimensional low sample size (HDLSS) data but come with the cost of limited expressivity.

\begin{figure*}[htb!] %
\vskip -0.12in 
    \centering
    {\includegraphics[width=\textwidth]{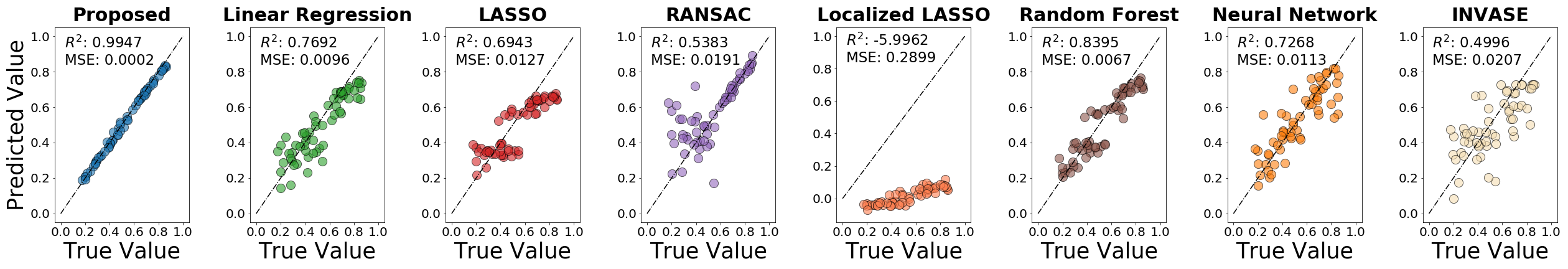}}%
    \caption{Strong ML models fail to learn the correct regression function when the target depends on different subsets of variables for different samples (see data model in Eq. \ref{eq:linear_regression_syn}).
    Each subplot presents the predicted (y-axis) vs. true (x-axis) values for different baselines. Points on the diagonal line indicate correct predictions. R-square and Mean Squared Error (MSE) are reported for each model. In this work, our proposed model (left plot) that effectively identifies the informative features for each sample while learning the regression coefficients. }
    \label{fig:linear_comp_more}%
\end{figure*}

In contrast, deep over-parametrized nets are highly expressive but tend to overfit on LSS tabular data; one reason is that in this regime, the vast amount of parameters leads to a large variance of the gradient estimates \cite{liu2017deep}. To prevent model overfitting, several authors \cite{DFS,sparseNN-group-lasso,one-layer,yamada2020feature} have proposed to apply different regularization schemes to sparsify the input features. Since these models select one global set of informative features, they are not suited for the heterogeneity of tabular biomedical data. These global feature selection models occasionally result in unsatisfactory performance and do not provide a sample-specific explanation for the predictions.

 Our working hypothesis is that since biomedical data is heterogeneous, different samples may require distinct prediction functions. Therefore, we design a simple yet remarkably effective NN-based framework that leads to dramatically higher prediction performance yet is intrinsically interpretable and is less prone to overfitting. Specifically, we propose a Locally SParse Interpretable Network (LSPIN), a NN that incorporates interpretability into its design (self-explanatory) by sparsifying the input variables used by the model locally (for each sample) while learning a \textit{prediction} function. To identify the local sparsity patterns, we train a \textit{gating} network to predict the probabilities of the instance-wise \textit{gates} being active. The parameters of the local gates, along with the model coefficients, are learned in tandem by minimizing a classification or regression loss. Our parametric construction leads to a highly interpretable (locally sparse) model which relies on a small subset of the input features for each instance. 

Our contributions are: (i) We propose a probabilistic $\ell_0$ like regularization that leads to sample-specific feature selection that is \textit{stable} across close samples defined by an affinity kernel. (ii) By training our \textit{gating} network alongside a \textit{prediction} network, we obtain a powerful interpretable NN framework for tabular biomedical data. (iii) We show via extensive synthetic simulations that our model, albeit simple, can learn the correct target function and identify the informative variables while requiring a small number of observations. (iv) We demonstrate a strong property of our framework: a linear predictor with local sparsity often outperforms state-of-the-art nonlinear models when applied to real-world datasets. (v) We explore the applicability of the proposed approach to several challenging tasks in biomedicine, including survival analysis and marker genes identification.

\section{Motivating Example}
\label{sec:mot_exp}

To motivate our proposed framework, let us consider a data matrix $\myvec{X}$, with $N$ measurements (e.g., patients) and $D$ variables (e.g., genes). Given a target variable $y$, in supervised learning, we are interested in modeling a function $\myvec{f}$ that can predict the target $y$ based on the observations $\myvec{X}$. Unfortunately, if $D>N$ learning such function becomes challenging and may lead to overfitting. Moreover, since the biomedical datasets are heterogeneous, the informative variables may vary within different population subsets. To clarify this point, let us consider the following simple linear regression problem. We are given a small ($N=10$) synthetic dataset in which the response variable ${y}$ of different samples depends on different subsets of features of the data matrix $\myvec{X}$. We assume that the data comprise two subpopulations, which we term here group-$1$ and group-$2$. The corresponding response variable ${y}$ for samples in group-$1$ and group-$2$ is defined in Eq. \ref{eq:linear_regression_syn} where for group-$1$, the response is a linear combination of the $1^\text{st}$,$2^\text{nd}$, and $3^\text{rd}$ features, and for group-$2$ it is a linear combination of the $3^\text{rd}$,$4^\text{th}$, and $5^\text{th}$ features:
\begin{align}
    {y} &= 
    \begin{cases}
        -2\myvec{x}_1 + \myvec{x}_2 - 0.5\myvec{x}_3 ,&\text{if in group 1,}\\ 
        -0.5\myvec{x}_3 + \myvec{x}_4 - 2\myvec{x}_5 ,& \text{if in group 2}.\\
    \end{cases}
    \label{eq:linear_regression_syn}
\end{align}
Group 1 and 2 are defined by drawing values for $\myvec{x}_1-\myvec{x}_5$ from separated Gaussians (details appear in Appendix section \ref{sec:linear_details}). The simple example above comprises two different linear relationships between the response $y$ and the observed variables. Since we do not know the membership of each point to one of the two groups, attempting to fit a single model to this data is challenging. In fact, in Fig. \ref{fig:linear_comp_more} we demonstrate that strong ML models fail to learn the correct regression function. In this study, we design LSPIN that is able to learn the correct target function while accurately identifying the informative features for each sample (see Appendix Fig. \ref{fig:lin_gate_com}).

\section{Problem Setup and Method}
\label{gen_inst}

We are interested in the standard supervised learning based on tabular biomedical data points $\{\myvec{x}^{(i)},y^{(i)}\}^N_{i=1}$, where $\myvec{x}^{(i)}\in \mathbb{R}^D$, with ${x}^{(i)}_d$ representing the $d^{th}$ feature of the $i^{th}$ vector-valued observations. Our goal is to design a method that can overcome the challenges posed by biomedical datasets while leading to \textit{accurate} and \textit{interpretable} predictions. Specifically, we want to learn an intrinsically interpretable prediction model $\myvec{f}_{\myvec{\theta}} \in \mathcal{F}$ with: \\{\bf P1} Small generalization error even in cases of $N<D$.\\ {\bf P2} Sample-specific removal of nuisance variables whose inclusion could be detrimental for predictions.\\
    {\bf P3} High expressive power. 
    
    Perhaps the most simple model that leads to {\bf P1} is the LASSO \cite{Lasso}. The LASSO minimizes the empirical risk of a linear model $\myvec{f}(\myvec{x}^{(i)})=\myvec{\theta}\myvec{x}^{(i)}$ (with $\myvec{\theta}^T\in \mathbb{R}^D)$, while penalizing for the sum of absolute values of active coefficients. This global linear feature selection model is interpretable since it provides an additive quantification to the contribution of each variable. To enable sample-specific variable selection ({\bf P2}), the Localized LASSO was introduced in \cite{yamada2017localized}. The authors introduce local weights $\myvec{\theta}^{(i)}$ to the following model $\myvec{f}(\myvec{x}^{(i)})=\myvec{\theta}^{(i)}\myvec{x}^{(i)}$, and minimize the empirical risk with a network type regularization $\lambda_1\sum r_{i,j} \|\myvec{\theta}^{(i)}-\myvec{\theta}^{(j)}\|_2+\lambda_2\sum\|\myvec{\theta}^{(i)}\|^2_1$. The first term regularizes models parameters to be similar if $r_{i,j}>0$, where the values of $r_{i,j}$ are given by a graph that represents affinities between samples. The second term encourages local sparsity. While the Localized LASSO addresses {\bf P1} and {\bf P2}, it only relies on linear relations between features to learn the coefficients $\myvec{\theta}^{(i)}$. Furthermore, the model has limited generalization capabilities since coefficients of unseen samples are estimated based on neighbors in the training set.

In this work, we extend the Localized LASSO by using a NN framework to learn the local sparsity patterns and enable more expressive prediction functions. To this end, we propose a NN framework with local sparsity such that predictions are only based on a small subset of features ${\cal S }^{(i)} \subset \{1,2,\hdots,D\},i=1,...,N$ which is optimized for each sample individually. By forcing $|{\cal S}^{(i)}| \ll D$, we can reduce the generalization gap of the model and use the (sample-specific) subset of selected features to interpret the prediction model.

\subsection{Locally Sparse Predictor}
 Given labeled observations $\{\myvec{x}^{(i)},y^{(i)}\}^N_{i=1}$, we want to learn a global prediction function $\myvec{f}_{\myvec{\theta}}$ (parametrized using a NN) and sets of indicator vectors $\myvec{s}^{(i)}\in \{0,1\}^D$ ($s^{(i)}_j = 1$ if $j \in {\cal S}^{(i)}$ and $0$ otherwise) that will "highlight" which subset of variables the model should rely on for the prediction of each target value $y_i$. This will enable the model to attain on less features for each sample and therefore reduce overfiting. 

Such a model can be learned by minimizing the following empirical regularized risk
\begin{equation}\label{eq:risk}
    \frac{1}{N}\sum^N_{i=1} {\cal{L}}\big(\myvec{f}_{\myvec{\theta}}(\myvec{x}^{(i)}\odot\myvec{s}^{(i)}),y^{(i)}\big) +\frac{\lambda}{N} \sum^N_{i=1}\|\myvec{s}^{(i)} \|_0,
\end{equation}
where $\cal{L}$ is a desired loss function (e.g., cross-entropy), and $\odot$ represents the Hadamard product (element-wise multiplication), and $\lambda$ is a regularization parameter that controls the sparsity level of the model. Unfortunately, due to the discrete nature of the $\ell_0$ regularizer, this objective is not differentiable, and finding the optimal solution becomes intractable. Moreover, even if it finds the optimal solution, it is not clear that the model would not overfit when $N<D$. Explicitly, in this regime, the model may select one feature for each sample and "memorize" the training set. Also, it is unclear how such a model can generalize on unseen samples (at test time), namely, because we need to predict the indicator vectors $\myvec{s}$ for the unlabeled data.

\subsection{Probabilistic Reformulation of the $\ell_0$ Regularizer}
Fortunately, the $\ell_0$ norm can be relaxed via a probabilistic differentiable counterpart. Specifically, by replacing the binary indicator vector $\myvec{s}$ with a Bernoulli vector $\mathbf{\tilde{s}}$, with independent entries which
satisfy $ \prob(\tilde{\textnormal{s}}_d = 1)=\pi_d$ for $d \in \{1,2,\hdots,D\}$. Such probabilistic formulation (of binary indicator vectors) converts the combinatorial search (over the discrete space of $\myvec{s}\in \{0,1\}^D$) to a search over the continuous space of Bernoulli parameters ($\myvec{\pi}\in [0,1]^D$). This formulation becomes useful in several applications such as: model compression \cite{Louizos2017LearningSN}, feature selection \cite{yamada2020feature,lindenbaum2021differentiable,shaham2021deep}, discrete softmax \cite{jang2016categorical}, sparse canonical correlation analysis \cite{lindenbaum2020deep} and many more.

By replacing the deterministic vectors $\myvec{{s}}^{(i)}$ in Eq. \ref{eq:risk} with their probabilistic counterparts $\mathbf{\tilde{s}}^{(i)}$ we can now differentiate through the random variables using REINFORCE \cite{reinforce} or REBAR \cite{tucker2017rebar}. However, these methods suffer from high variance and require many Monte Carlo samples. Alternatively, as demonstrated in \cite{maddison2016concrete,yamada2020feature} a continuous reparametrization of the discrete random variables can reduce the variance of the gradient estimates. In the next section, we propose to learn the indicator vectors $\myvec{s}^{(i)}(\myvec{x}_i)$ by re-formalizing them as random vectors whose parameters (probabilities of being active) are predicted using a NN.

\begin{figure*}[htb!] 

\centering

\includegraphics[width=.95\textwidth]{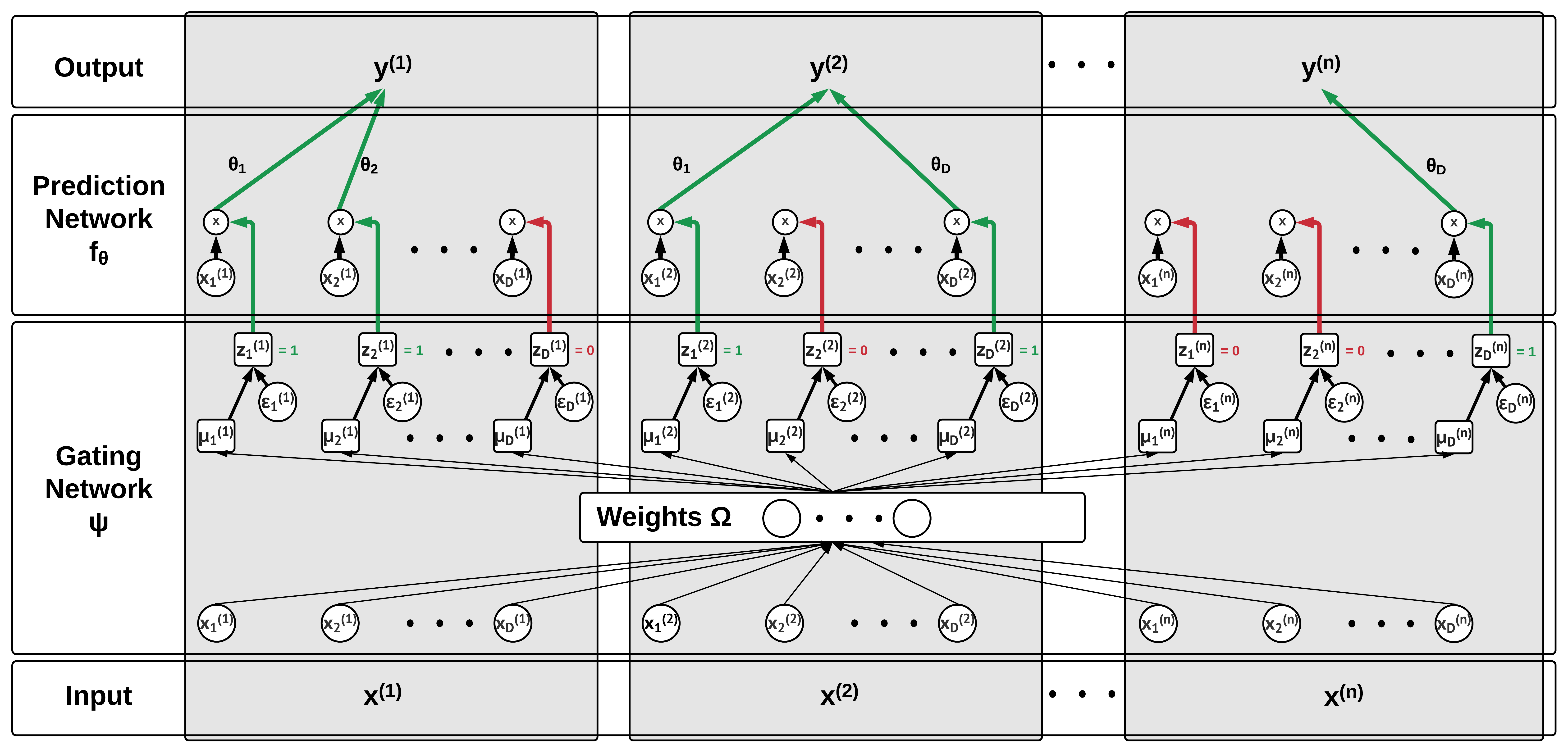}  

\caption{The architecture of Locally Linear SParse Interpretable Networks (LLSPIN). The data $\{\myvec{x}^{(i)} = [x_1^{(i)},x_2^{(i)},...,x_D^{(i)}]\}^n_{i=1}$ is fed simultaneously to a \textit{gating} network $\myvec{\Psi}$ and to a \textit{prediction} network $\myvec{f}_{\myvec{\theta}}$ (which is linear in this example). The \textit{gating} network $\myvec{\Psi}$ learns to predict a set of parameters $\{\mu_d^
{(i)}\}^{D,n}_{d=1,i=1}$. The parameters $\mu_d^{(i)}$ depict the behavior of local stochastic gates $\textnormal{z}_d^{(i)}\in [0,1]$ that sparsify (for each instance $i$) the set of features that propagate into in the prediction model $\myvec{f}_{\myvec{\theta}}$. LLSPIN leads to sample-specific (local) sparsification (obtained via the \textit{gating} network). Therefore, it can handle extreme cases of LSS and lead to interpretable predictions (since the model only uses a small subset of features for each sample). For illustration purposes, we overlay this figure using green (active) and red (non-active) arrows, which indicate that some samples require two features while others only one, in this example. In Section \ref{sec:experiments}, we demonstrate using extensive experiments that our model, leads to accurate predictions in the challenging regime of LSS data.  }
\label{fig:LLSPIN_diagram}
\end{figure*}

\subsection{Locally Sparse NN}
\label{sec:cr}
This section describes the proposed Locally SParse Interpretable Network (LSPIN)\footnote{Codes are available at https://github.com/jcyang34/lspin}. LSPIN is a \textit{prediction} NN with sample-specific gates which sparsify the variables used by the model locally. The sparsity patterns of the gates are learned via a second NN. This leads to a natural framework to predict the sparsity patterns of unseen samples (see illustration in Fig. \ref{fig:LLSPIN_diagram}).

Each stochastic gate (for feature $d$ and sample $i$) is defined based on the following hard thresholding function 
$$\textnormal{z}^{(i)}_d = \max(0,
\min(1,0.5+ \mu^{(i)}_d + \epsilon^{(i)}_d)),$$ where ${\epsilon}^{(i)}_d$ is drawn from  $\mathcal{N}(0 ,\sigma^2)$ and $\sigma$ is fixed throughout training. The choice of $\sigma$ (which controls the injected noise) is discussed in Section \ref{sec:reproduce} in the Appendix. 
The sample-specific parameters ${\myvec{\mu}^{(i)}}\in \mathbb{R}^D,i=1,...,N$ are predicted based on a \textit{gating} network $\myvec{\Psi}$ such that ${\myvec{\mu}^{(i)}}=\myvec{\psi}(\myvec{x}^{(i)}|\myvec{\Omega})$, where $\myvec{\Omega}$ are the weights of the \textit{gating} network. These weights are learned simultaneously with the weights of the \textit{prediction} network by minimizing the following loss:
 
\begin{equation}
\label{eq:loss}
\mathbb{E}\big[ {\cal{L}} (\myvec{f}_{\myvec{\theta}}(\myvec{x}^{(i)}\odot\mathbf{z}^{(i)}),y^{(i)}) +{\cal{R}}(\mathbf{z}^{(i)}) \big],
\end{equation}
where $\cal{L}$ is a desired loss (e.g. cross entropy), and we compute its empirical expectation over $\myvec{x}^{(i)},y^{(i)}$ and $\mathbf{z}^{(i)}$, for $i$ in a batch of size $B$. The term ${\cal{R}}(\mathbf{z}^{(i)})$ is a regularizer that we define as
\begin{equation}\label{eq:reg}
   {\cal{R}}(\mathbf{z}^{(i)})=\lambda_1 \|\mathbf{z}^{(i)} \|_0+\lambda_2\sum_j K_{i,j}\| \mathbf{z}^{(i)}- \mathbf{z}^{(j)}\|_2^2.
\end{equation}
After taking the expectation (over $\mathbf{z}^{(i)}$), the leading term in $\cal{R}$  can be rewritten using a double sum in terms of the Gaussian error function ($\erf$): 


\begin{equation*}
    \frac{1}{N}\sum^N_{i=1} \sum^D_{d=1}\left(\frac{1}{2} - \frac{1}{2} \erf\left(-\frac{\mu^{(i)}_d + 0.5}{\sqrt{2}\sigma}\right) \right).
\end{equation*}
The second regularization term (in \ref{eq:reg}) is introduced to encourage {\bf stability} of the local variable selection mechanism and is evaluated using Monte Carlo sampling. The kernel $K_{i,j}\geq 0$ is user defined (e.g. radial basis function) and should reflect the affinity between samples $\myvec{x}^{(i)}$ and $\myvec{x}^{(j)}$, therefore, we can ensure that for nearby points our model would lead to similar sparsity patterns in $\mathbf{z}^{(i)}$ and $\mathbf{z}^{(j)}$.
Altogether, Eq. \ref{eq:loss} is optimized using SGD over the model parameters $\myvec{\theta}$ and the parameters of the \textit{gating} network $\myvec{\Omega}$ (see Algorithm \ref{alg:pseudocode} in the Appendix for description of training procedure). If $\mu^{(i)}_d$ of sample $i$ is a large number, then the $d$-th feature will be relevant for predicting $y^{(i)}$ with high probability (and vice versa for very small numbers). The stochasticity of the model plays two important roles: (1) it allows us to train weights of a binary model (the \textit{gating} network). (2) it enables the model to re-evaluate features that are sparsified at an early step of training.

\begin{table*}[htb!]
    \centering
    \begin{adjustbox}{max width=1 \textwidth,valign=c}
    \begin{tabular}{|c|| c|c|| c|c|| c|c|| c|c||c|c||}
    \hline
    & \multicolumn{2}{c||}{\textbf{E1}} & \multicolumn{2}{c||}{\textbf{E2}}  & 
    \multicolumn{2}{c||}{\textbf{E3}} & 
    \multicolumn{2}{c||}{\textbf{E4}} &
    \multicolumn{2}{c||}{\textbf{E5}} \\
    \hline
    & F1 & ACC & F1 & ACC & F1 & ACC & F1 & ACC & F1 & MSE\\
    \hline
    
    LASSO & 0.5000 & 52.00 & 0.5000 & 74.50 &  0.6250 & 71.50 & 0.1290 & 64.00 & 0.3704 & 1.0190\\
      
      RF & 0.5333 & 88.50 & 0.5333 & 88.50 & 0.6250 & 87.00 & 0.0769 & 86.00 & 0.2500 & 0.2499 \\
      
      INVASE & 0.5390 & 89.00 & 0.7000 & 88.00 & 0.6923 & 86.00 & 0.6667 & 94.00 & 0.1526 & 3.1264\\

      L2X & 0.7986 & 88.00 & 0.6050 & 94.50 & 0.2450 & 87.00 & 0.5000 & 92.00 & 0.6081 & 0.5134\\ 
      
      
      
      TabNet & 0.4789 & 54.50 & 0.5426 & 65.50 & 0.6905 & 78.50 & 0.0036 & 60.00 & 0.4454 & 1.0317 \\
      
      REAL-x & 0.8306 & 85.00 & 0.7089 & 88.50 & 0.7823 & 86.00 & 0.8511 & 90.00 & NA* & NA* \\ 
    
    LLSPIN & 0.3337 & 80.50 & 0.7216 & 86.50 & 0.4741 & 
    73.50 & 0.9458 &
    90.00 & 0.6815 & 0.4927\\
    
    
    LSPIN & \textbf{0.9761} & \textbf{94.00} & \textbf{0.8600} & \textbf{95.00} & \textbf{0.9296} & \textbf{89.00} & \textbf{0.9615} & \textbf{98.00} & \textbf{1} & \textbf{0.0019} \\

      \hline
    \end{tabular}
    \end{adjustbox}
    \caption{Nonlinear synthetic datasets (see Eqs. \ref{eq:exp1}-\ref{eq:exp5} in Appendix section \ref{sec:nonl_details}). We compare the proposed LLSPIN/LSPIN to other baselines in terms of the F1 score of the selected features and accuracy($\%$)/ MSE for the prediction performance. Across these examples, LSPIN correctly identifies the informative features with a substantially higher F1 and higher accuracy/lower mse. *We attempted to implement REAL-x for the regression task, but the model failed to converge.}
    \label{tab:nonl_metric_comp}
\end{table*}
For inference, we remove the stochasticity from the gates and set $\hat{z}^{(i)}_d = \max(0,\\
\min(1,0.5+ \mu^{(i)}_d))$, which informs what features are selected. In practice, we observe that the coordinates of $\hat{z}^{(i)}_d$ mostly converge to $0$ or $1$ (see statistics in Table \ref{tab:gate_convergence} in the Appendix). This solution is encouraged as it is stable to the injected Gaussian noise ($\epsilon^{(i)}_d$). Namely, once $\mu^{(i)}_d=\{-1,1\}$ (which are at the boundary of the range of the tanh activation used in the \textit{gating} network), the value of the corresponding $\textnormal{z}^{(i)}_d$ would be with high probability $0$ and $1$, respectively. This is because the injected noise is less likely to push the values of $\textnormal{z}^{(i)}_d$ into the range $(0,1)$. 

\section{Related Work}

Identifying sample-specific subsets of variables that are important for prediction has been studied in the context of interpretability. Methods such as \cite{simonyan2013deep,zeiler2014visualizing,lundberg2017unified} try to identify a small subset of features that explain the predictions made by a pre-trained model. These models either use the gradients of the pre-trained model or use perturbations to study the influence of different variables on the predictions of each instance. However, as shown in \cite{jethani2021have}, these models either require heavy post-training computations or are inaccurate \cite{adebayo2018sanity,gale2019state}. More recent works such as \cite{dabkowski2017real,chen2018learning,schwab2019cxplain,yoon2018invase} alleviate the computational burden by training a single model to explain all samples. Still, they are all designed to explain pre-trained black-box models and thus cannot reduce the generalization gap in the case of LSS data.

Two recent works \cite{yoon2018invase,jethani2021have} present solutions which allow for training a prediction model in tandem with an explanatory model. However, both methods try to learn a model that ``imitates'' the predictions made by a baseline model, which uses the complete set of features. We argue that using the entire feature space in LSS data can lead to overfitting. Moreover, these methods require training a large number of parameters and use REINFORCE \cite{reinforce} or REBAR \cite{tucker2017rebar} for learning the sparsity patterns. We demonstrate in the benchmark experiment (see Fig. \ref{fig:time_benchmarking} in the Appendix) that the method proposed in \cite{yoon2018invase} is computationally expensive and does not scale well to large datasets. Furthermore, in Section \ref{sec:experiments} we provided extensive empirical evidence that our framework is more accurate and interpretable compared to \cite{yoon2018invase,jethani2021have}.

Sparsification of neural networks has also been utilized for other purposes, e.g., in Mixture of Experts \cite{shazeer2017outrageously,riquelme2021scaling} or for accelerating the inference of over parametrized deep nets, \cite{dong2017more,gao2018dynamic,ashouri2019retraining,kurtz2020inducing,fedus2021switch}. In contrast to these models, our method sparsifies the number of features used for predictions of each instance, thus leading to a more robust and interpretable model.

\section{Experiments}
\label{sec:experiments}

In this section, we evaluate how \textit{accurate} and \textit{interpretable} the proposed approach is on both synthetic and real-world biomedical datasets. We compare to: embedded feature selection methods such as LASSO \cite{Lasso}, linear support vector classification (SVC) \cite{SVC}, tree-based wrapper methods such as Random Forest (RF) \cite{Rf1} and XGBoost \cite{xgboost}, and neural network-based methods such as fully connected neural network (with no convolution, distortion, weight decay, or unsupervised pre-training), STG \cite{yamada2020feature}, INVASE \cite{yoon2018invase}, L2X \cite{chen2018learning}, TabNet \cite{arik2020tabnet}, and REAL-x \cite{jethani2021have}. Additionally we compare to RANSAC \cite{RANSAC} and Localized LASSO \cite{yamada2017localized} for the motivating example, and DeepSurv \cite{Jared}, COX-LASSO, COX-STG \cite{yamada2020feature}, and Random Survival Forest \cite{RSF} for the survival analysis result. We failed to compare to Localized LASSO, since is not suited for classification and it did not converge on the regression examples. Details of the datasets availability, training procedure, and hyper-parameter tuning are included in Appendix section \ref{sec:reproduce}.

\subsection{Nonlinear Prediction on Synthetic Datasets}\label{sec:synt}
This section uses synthetic datasets where the target value only depends on a subset of variables that varies across samples. Since the per sample subset of informative variables is known, we can perform a controlled evaluation of the predictivity and interpretability of our model. First, we focus on classification data models \textbf{E1}-\textbf{E3} which were also used for evaluation in \cite{yoon2018invase,jethani2021have,arik2020tabnet}. We further design a higher dimensional example \textbf{E4}, and a highly nonlinear "moving-XOR" regression example \textbf{E5}. In all data sets, we use less than $2000$ samples for training, a regime which is more challenging than what was previously studied by \cite{yoon2018invase,jethani2021have,arik2020tabnet}. The exact data models and training procedures are described in the Appendix section \ref{sec:nonl_details}. We evaluate all models by measuring the F1 score\footnote{$\text{F1} = \frac{\text{TP}}{\text{TP}+\frac{1}{2}(\text{FP}+\text{FN})}$ where $\text{TP}$ is the number of informative features that are selected by the model, $\text{FP}$ is the number of selected features that are uninformative, and $\text{FN}$ is the number of informative features unrecovered by the model.} of the selected features and the prediction performance (accuracy for classification and Mean Squared Error (MSE) for regression). 

As shown in Table \ref{tab:nonl_metric_comp}, LSPIN consistently outperforms existing baselines in terms of its ability to identify the informative variables (evaluated using the F1 score). At the same time, our model leads to improved predictive capabilities compared to these baselines. Our linear model (LLSPIN) leads to relatively high F1 score in \textbf{E2}, \textbf{E4}, \textbf{E5} despite the fact that the data contains nonlinear feature interactions.

\begin{figure*}[htb!]%
    \centering
    \includegraphics[width=0.98 \textwidth,valign=c]{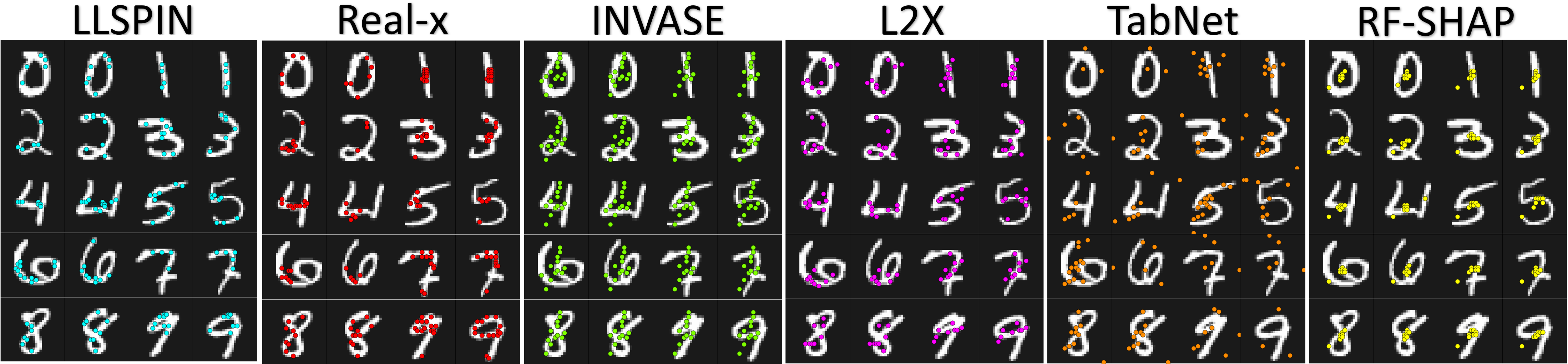}
    { \adjustbox{max width=1 \textwidth,min height = 0.6 in, valign=c}{
             \begin{tabular}{|l|l|c|c|c|c|c|c|}
                \hline 
              Method & ACC & $\#$ Feat & Stability & Diversity &  Faithfulness & Gen-SVM & Gen-$k$-means  \\
                 \hline 
                RF-SHAP & 96.81 & 8  & 0.418 & 41.15 &0.531 & 52.42 & 40.15 \\
              REAL-x & 96.95 & 10  & 0.415 & 80.55 &0.885 & 94.04& 87.94 \\
              L2X   & 89.11 & 8 & 0.268 & 94.79 & 0.791 & 94.18 & 89.56\\
              INVASE & 85.07 & 11 & 0.162& 13.43 & 0.864& 69.67 & 43.02\\
              TabNet & 96.79 & 6 & 0.265& 89.17 & 0.759 & 54.42 & 43.22 \\
              LLSPIN ($\lambda_2=0)$ & 98.26 & 7 & 0.294 & 99.14& 0.950 & 98.22 & 97.50 \\
              LLSPIN ($\lambda_2=0.1)$ & 98.18 & 6 & 0.098 & 99.05& 0.926 & 96.23 & 96.84 \\
              LSPIN ($\lambda_2=0)$ & \textbf{98.45} & 7 & 0.256 & \textbf{99.84} & \textbf{0.987} & \textbf{98.43} & \textbf{97.99} \\
           LSPIN ($\lambda_2=0.1)$ & 98.29 & 6 & \textbf{0.065} & 99.39 & 0.917 & 98.42 & 97.57 \\
                 \hline 
         \end{tabular}%
  \label{tab:addlabel}%
    }
    }%
    \caption{Top: Random samples from the MNIST dataset. For each example, we overlay the image with color dots indicating locations of the estimated informative pixels by each method. Bottom: Test accuracy (ACC), the median number of features selected by each the model ($\#$ Feat), and several metrics for evaluating interpretability capabilities (explained in Section \ref{sec:mnist_exp}). Our model leads to high classification accuracy while relying on a few faithful and generalizable features. The selected features by our model are diverse across different classes while remaining stable for nearby samples. We compare our models with and without the second regularization term ($\lambda_2$ in Eq. \ref{eq:reg}), and demonstrate that it improves stability.}%
    \label{fig:MNIST-gates}%
\end{figure*}

\subsection{Interpretability Evaluation on MNIST}
\label{sec:mnist_exp}
In this section, we demonstrate that the proposed method is exceptionally \textit{interpretable} while leading to predictions that are more \textit{accurate} than state-of-the-art non-linear models. To evaluate interpretability, previous authors suggest the following criteria:\\ 
    {\bf Faithfulness: } Are the identified features significant for prediction?\\
    {\bf Stability: } Are explanations to similar samples consistent?\\
     {\bf Diversity:} How different are the selected variables for instances of distinct classes?\\
    {\bf Generalizability:} Are the selected features beneficial for making accurate predictions using other simple models? 

We use MNIST handwritten dataset as a table with 784 features and do not consider spatial information (since we are interested in tabular data). We compare our model to RF with SHAP \cite{lundberg2017unified} and leading NN interpretability models, including L2X, REAL-x, INVASE, and TabNet.

As suggested by \cite{alvarez2018towards}, {\bf faithfulness} could be evaluated by removing features one by one (based on their importance) and calculating the correlation between the predictivity drop and the feature importance. Following \cite{alvarez2018towards,yoshikawa2020gaussian}, {\bf stability} is evaluated by computing the Lipchitz constant of the explanation function. This is estimated for $\myvec{x}_i$ using $\underset{\myvec{x}_i,\myvec{x}_k\leq {\epsilon}}{\max} \frac{\|\myvec{w}_i-\myvec{w}_k \|_2}{ \|\myvec{x}_i-\myvec{x}_k \|_2}$, where $\myvec{w}_i$ is the explanation vector for sample $i$ provided by each method. Then we average the Lipchitz constant over all samples. To evaluate {\bf diversity} we take the per class median selected features and use the Jaccard index to count the portion of non-overlapping selected features across classes (exact formula appears in Appendix \ref{sec:diverse}). To evaluate the {\bf generalizability} of the selected features, we measure the accuracies of SVM and $k$-means when applied to the data, which is masked by the selected features. We expect the performance to be preserved if the selected features are crucial for prediction (compared to the accuracy obtained when using all features).

We tune all models to identify the $\sim 10$ most informative pixels per image and present these in Fig. \ref{fig:MNIST-gates}. As visually indicated by this figure, LLSPIN, Real-X, and L2X tend to select pixels with non-zero values that cover ``unique'' patterns in the digits. INVASE seems to lead to a more global selection; RF-SHAP concentrates on one region per digit, and TabNet selects several non-active pixels. Since TabNet still leads to high classification accuracy, we suspect that it encodes predictions in its interpretation (as suggested in \cite{jethani2021have}).

Next, we use the interpretability metrics described above to compare all baselines. As indicated in the table (see bottom panel of Fig. \ref{fig:MNIST-gates}), both LSPIN and LLSPIN perform exceptionally well in terms of accuracy and interpretability. We compare our models with and without the second regularization term (see $\lambda_2$ in Eq. \ref{eq:reg}). Our results suggest that including this term improves the stability of the selected features without a significant compromise for other qualities. The results highlight three encouraging findings: (i) our linear model leads to an accuracy close to CNN level (which is $98.9\%$). (ii) applying $k$-means to the data gates by the selected features leads to a dramatic improvement in clustering accuracy. Namely, it improves from $\sim 55\%$ to an accuracy higher than $96\%$ when using the features selected by LLSPIN or LSPIN. (iii) our model improves robustness and uncertainty estimates \cite{ovadia2019can} under distributional shifts compared with a standard fully connected network (see results in Appendix \ref{sec:shift}).

\subsection{Classification of LSS Real World Data}
\label{sec:real_world}
In this section, we evaluate LLSPIN and LSPIN on several challenging LSS real-world biomedical datasets (properties are summarized at the bottom of Table \ref{tab:real_data}). BASEHOCK, RELATHE, PCMAC, COLON, TOX171 are from the feature selection dataset collections\footnote{https://jundongl.github.io/scikit-feature/datasets.html}, and the purified PBMC dataset is from \cite{pbmcP}). To optimize the hyperparameters of all baselines, we applied a nested cross-validation procedure (see details in Appendix section \ref{sec:real_world_details}).   


In Table \ref{tab:real_data}, we present the average test accuracy and number of selected features for all baselines. As evident across several datasets, our framework dramatically improves the accuracy compared to standard MLP while using a small portion of the input set of variables. Moreover, in most cases, our model outperforms state-of-the-art models such as XGBoost, TabNet, and REAL-X. We rank the methods based on the average classification accuracy for each dataset. Our models (LLSPIN/LSPIN) reach the top 2 places based on the median rank across all datasets.

In these examples, to our surprise, LLSPIN outperforms LSPIN. We reason that LLSPIN (the linear version of our model) remains highly expressive in the high dimensional setting since it learns several linear relationships, each based on a small set of coefficients. Moreover, since this prediction model does not contain any nonlinearity, overfitting is less likely to happen than in the nonlinear models. These results suggest that LLSPIN can serve as an accurate and highly interpretable model in LSS data regimes.
In Appendix \ref{sec:shift} we evaluate our models predictive \textit{uncertainty} using the Negative Log Likelihood (NLL) and demonstrate that it leads to more calibrated uncertainty estimates \cite{ovadia2019can} compared with other baselines.

\begin{table*}[htb!]
    \centering
    \begin{adjustbox}{max width=1 \textwidth,min height = 0.6 in, valign=l}
    \begin{tabular}{|c|cccccc|c|}
    
    \hline

    & BASEHOCK & RELATHE & PCMAC & PBMC & COLON & TOX-171& Median Rank \\
 
         \hline
    
    LASSO &  $74.46\pm5.19\ [34]$  & $58.69\pm1.59\ [18]$ & $68.09\pm4.08\ [21]$ &  $90.30\pm0.36\ [31]$ &  $81.54\pm9.85\ [24]$ &   $ 87.71\pm4.62\ [49]$ & $6.5$ \\ 
         
    
    SVC  & $74.46\pm3.37\ [22]$ &  $56.48\pm3.00\ [6]$ & $67.41\pm3.72\ [12]$ & $89.02\pm0.74\ [30]$ & $76.15\pm9.39\ [25]$ & $81.14\pm7.47\ [38]$ & $8.5$\\
    
    RF  & $64.46\pm4.52\ [10]$ &  $71.42\pm3.50\ [50]$ & $67.44\pm7.00\ [9]$ & $48.56\pm6.18\ [10]$ & $79.23\pm9.76\ [47]$ & $53.71\pm9.96\ [42] $ & $11.5$ \\
    
    XGBoost &  $90.37\pm1.05\ [45]$ & $76.75\pm1.67\ [32]$ & $83.93\pm0.67\ [43]$ & $76.58\pm0.72\ [64]$ & $76.15\pm12.14\ [7]$ & $67.43\pm5.60\ [38]$ & $6$ \\
    
    MLP & $56.51\pm1.43$ & $55.44\pm2.38$ & $54.38\pm1.27$ & $61.57\pm1.45$ & $81.54\pm7.84$ & $62.59\pm8.03$ & $12.5$\\
    
    Linear STG &  $89.36\pm1.40\ [27]$ & $69.94\pm5.05\ [16]$ & $\bm{85.11\pm1.07\ [42]}$ & $88.22\pm0.82\ [27]$ & $74.62\pm11.44\ [14]$ & $71.14\pm5.78\ [16]$ & $7$ \\
    
    Nonlinear STG & $89.24\pm1.18\ [20]$ & $74.83\pm3.95\ [27]$ & $84.16\pm0.90\ [32]$ & $86.29\pm1.31\ [19]$ & $76.15\pm13.95\ [8]$ & $67.43\pm7.25\ [14]$ & $6.5$\\

    INVASE & $84.02\pm0.81\ [42]$ & $70.81\pm1.56\ [43]$ & $77.06\pm1.01\ [48]$ & $86.34\pm0.81\ [30] $ & $76.92\pm12.40\ [6]$ & $76.86 \pm 7.39 [26]$ & $7.5$ \\

    L2X & $88.48\pm2.01\ [1]$ & $77.10\pm5.19\ [10]$ & $78.69\pm3.62\ [10]$ & $70.77\pm11.24\ [10]$ & $78.46\pm8.28\ [8]$& $71.71\pm10.42 [9]$ & $6.5$ \\ 
    
    TabNet & $88.21\pm2.00\ [3]$ & $67.84\pm15.40\ [10]$ & $69.35\pm10.49\ [4]$ & $\bm{92.13\pm0.59\ [3]}$ & $64.62\pm12.02\ [28]$& $30.00\pm6.29\ [34]$ & $9.5$\\
    
    REAL-x & $89.80\pm1.96\ [5]$ & $80.61\pm1.31\ [3]$& $80.98\pm3.05\ [6]$ & $83.39\pm2.19\ [24]$& $75.38\pm12.78\ [15]$& $77.71\pm7.65\ [42]$ & $5$\\
    
    LSPIN & $89.37\pm1.48\ [3]$ & $80.59\pm1.95\ [3]$ & $78.51\pm1.48\ [3]$ & $88.67\pm0.64\ [15]$ & $71.54\pm6.92\ [1]$ & $90.29\pm5.45 [1]$  & $4.5$\\
    
    LLSPIN & $\bm{91.56\pm1.51\ [4]}$& $\bm{82.01\pm 2.20\ [11]}$ & $81.48\pm1.74\ [3]$ & $90.43\pm0.6\ [18]$ & $\bm{83.85\pm 5.38\ [7]}$ & $\bm{92.57\pm6.41\ [6]}$ & $\bm{1}$\\
    
    \hline
    Train\ /$\ $Test\ & $379\ $/$\ 1514\ $ &
    $271\ $/$\ 1084\ $& $\ 369\ $/$\ 1476\ $& $\ 721\ $/$\ 2880\ $& $\ 49\ $/$\ 13\ $& $\ 136\ $/$\ 35\ $\\\cline{1-7}

   $\ $Dim/\ Classes & $\ 4862\ $/$\ 2\ $ & $\ 4322\ $/$\ 2\ $ & $\ 3289\ $/$\ 2\ $ & $\ 2000\ $/$\ 4\ $ & $\ 2000\ $/$\ 2\ $ & $\ 5748\ $/$\ 4\ $ \\\cline{1-7}

   \end{tabular}
    \end{adjustbox}
    \caption{Classification on biomedical tabular datasets. We report the average accuracy and standard deviation, with the corresponding median number of selected features in square brackets. The number of training/test samples, dimensions, and classes are also reported.}
    \label{tab:real_data}
\end{table*}

\begin{figure*}[htb!]%
    \centering
    {\includegraphics[width=0.4 \textwidth]{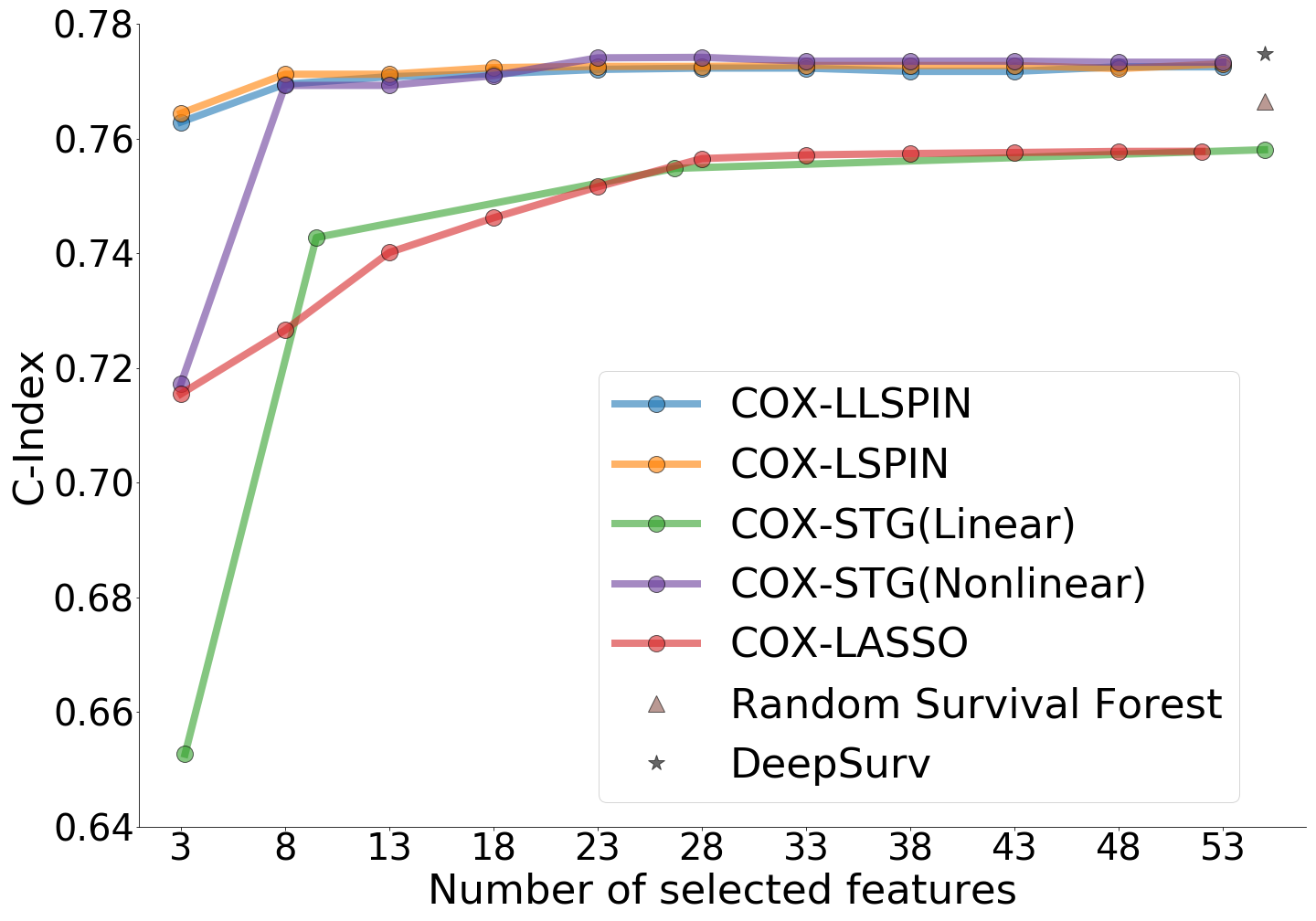}}
    {\includegraphics[width=0.4 \textwidth]{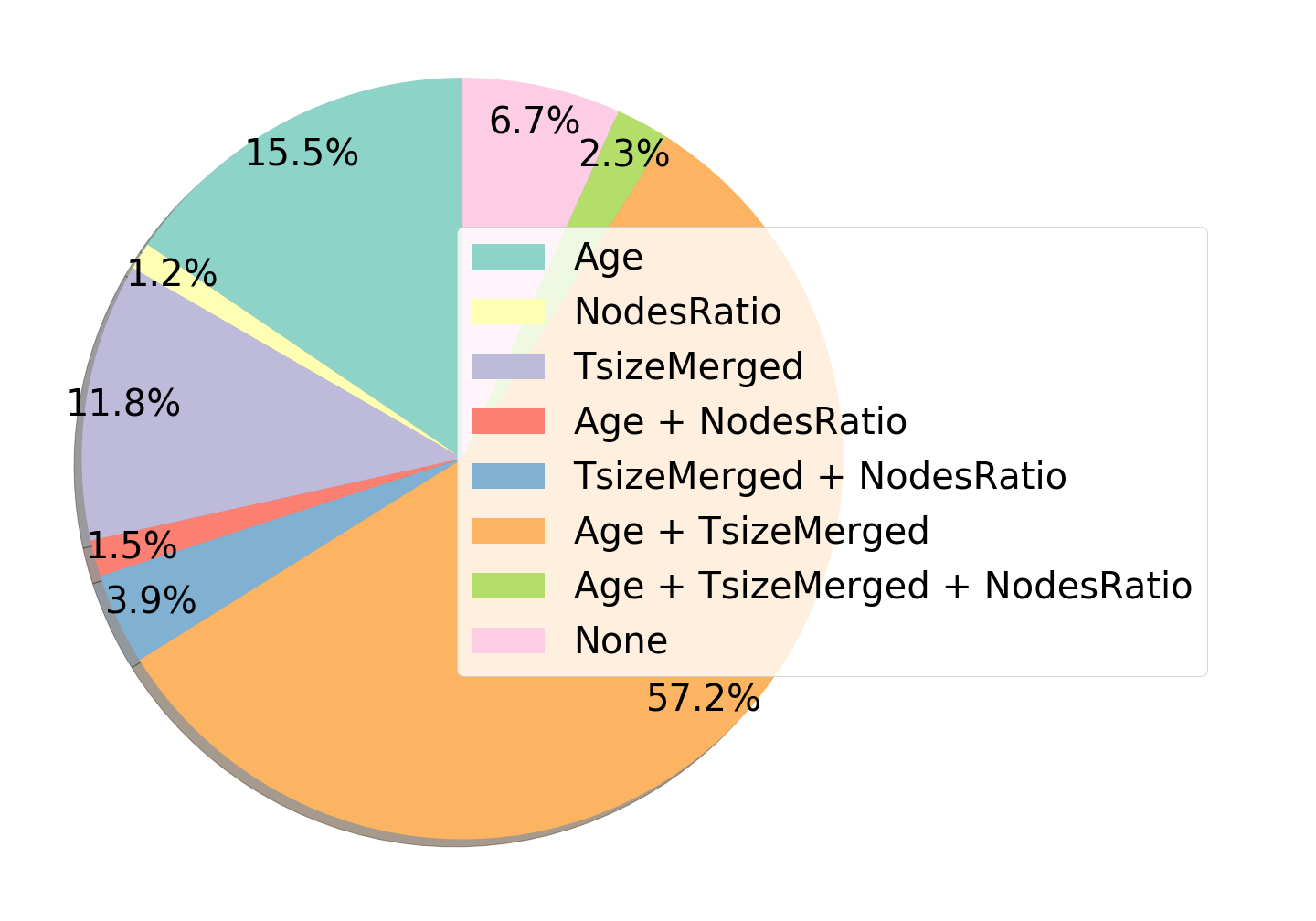}}
    \caption{Survival analysis based on the SEER breast cancer data. Left: Comparing the test C-Index obtained using subsets of most informative features. Right: Frequency of variables selected by COX-LLSPIN across the different patients. }%
    \label{fig:cox_result}%
\end{figure*}

\subsection{Survival Analysis}  \label{sec:cox}
Survival Analysis involves predicting the survival time of individual patients based on different clinical variables. In Survival Analysis, instance-level interpretation of the selected features is of particular interest as it can answer what are variables have the most significant effect on the survival of individual patients. We integrate our models (LSPIN/LLSPIN) into DeepSurv \cite{Jared}, which is a neural network framework for Cox regression. Then, we apply the integrated models (COX-LLSPIN/COX-LSPIN) on a Surveillance, Epidemiology, and End Results (SEER) breast cancer dataset \cite{national2011surveillance} to perform survival analysis.

We evaluate the performance of our models by computing test Concordance Index (C-Index) w.r.t. the number of selected features. We compare the performance with other Survival Analysis models as shown in Fig. \ref{fig:cox_result} (Left). We can see that the COX-LLSPIN/COX-LSPIN is comparable to state-of-the-art schemes when using more than $28$ features and outperforms all existing methods when focusing on small subsets of selected features. More importantly, our model (COX-LLSPIN) can provide more interpretable results while remaining accurate. 


Fig. \ref{fig:cox_result} (Right) shows the frequency of the selected feature sets among the different subjects (samples). For instance, $57.2\%$ of the samples have Age and TsizeMerged selected as important factors for the prediction. In contrast, $1.2\%$ of the samples have only NodesRatio selected, demonstrating that our models can characterize the heterogeneity among these samples. We argue that this is an important property for practitioners since knowing what variables affect each patient's outcome can improve personalized treatments.

\subsection{Marker Gene Identification}
\label{sec:sn_results}

Accurate cell classification is imperative for the success of many single-cell genomics studies. Developing an automated way to identify genes that allow identification of cell types (marker genes) is an ongoing challenge \cite{dai2021accurate}.
Here, we apply our model to a Single Nucleus RNA-sequencing dataset. The cell types in this data, namely Microglia and Oligodendrocyte Precursor Cells, are well characterized by \textit{ITGAM} gene and \textit{PDGFRA} gene, respectively. We aim to apply our model to identify these markers for each cell type automatically. Details of the data preprocessing and split are in Appendix section \ref{sec:snrna_details}.

Towards this goal, we aim to encourage our model to select a diverse set of features for each class. Therefore, we modify the second term in our regularizer (see Eq. \ref{eq:reg}) to $\lambda_2\sum_{j} (1-K_{i,j})\times(-\| \mathbf{z}^{(i)}- \mathbf{z}^{(j)}\|_2^2)$. Intuitively, when sample $i$ and $j$ are dissimilar ($K_{i,j}$ is small), the corresponding gates are encouraged to be different. Here, we fix $\lambda_2=1$. 

LLSPIN successfully identifies the two cell-type-specific markers (see Appendix Fig. \ref{fig:cell_type_markers}) while predicting the correct cell with $99.0\%$ accuracy. We further evaluate other instance-wise feature selection methods on this example. As indicated by the F1 score of the selected genes in Table \ref{tab:sc_f1} our approach significantly outperforms other schemes in its ability to identify the marker genes correctly.

\begin{table}[htb!]
    \centering
    \begin{adjustbox}{max width=0.8 \textwidth,max height = 0.4 in,valign=c}
    \begin{tabular}{|c|c|c|c|c|c|}
    \hline
         &  LLSPIN & INVASE & L2X & TabNet & REAL-x\\
    \hline
         F1 & $\bm{0.9950}$ & $0.4900$ & $0.4900$ & $0.2817$ & $0.5000$\\
         \hline
    \end{tabular}
    \end{adjustbox}
    \caption{Marker gene identification using several baselines. LLSPIN accurately identifies the known marker genes reflected by the F1 score computed based on selected features.}
    \label{tab:sc_f1}
\end{table}


\section{Conclusion}
We present a NN framework for making \textit{accurate} and \textit{intepratable} predictions based on tabular biomedical detests. To achieve these goals, we design a special kind of sample-specific regularizer that leads to sparsification that is \textit{stable} for similar samples. Our regularizer is parametrized using a \textit{gating} network that is trained simultaneously with a \textit{prediction} network and learns for each sample the set of most informative features. This leads to an intrinsically interpretable model, which can handle cases of low-sample-size (LSS) data that is either high dimensional or contains nuisance features. We demonstrate using synthetic and real datasets that our model can outperform state-of-the-art classification and regression models. Furthermore, when applied to datasets with nuisance variables, our model correctly identifies the subsets of informative features.

\section*{Acknowledgements}

The authors thank Mihir Khunte and Michal Marczyk for the preprocessing steps of the SEER breast cancer data.

\bibliography{ref}
\bibliographystyle{unsrt}

\newpage
\appendix
\renewcommand\thefigure{\thesection.\arabic{figure}} 
\setcounter{figure}{0}  
\renewcommand\thetable{\thesection.\arabic{table}} 
\setcounter{table}{0}  

\onecolumn
\section{Additional Results}\label{sec:result_additional}
In the following sections, we provide additional experimental results to support the effectiveness of the proposed approach.

\subsection{Visualization of the Motivating Example and Extended Evaluations}
\label{sec:linear_varying}
\begin{figure}[htb!] %
    \centering
    
    \includegraphics[width=0.35\textwidth]{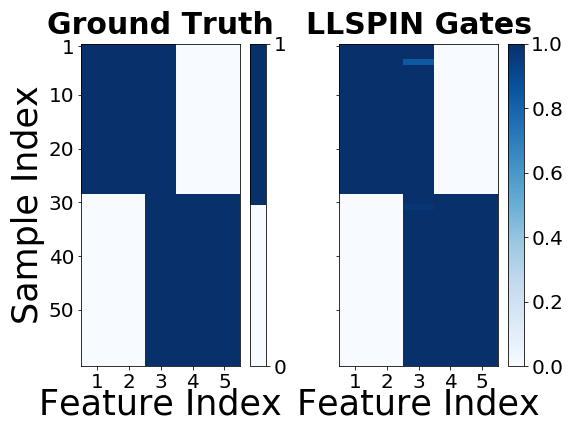}
    
    \caption{Heat-map comparison between the ground truth informative features for each sample (Left, $1$ for truly explanatory and $0$ for not) and the identified features by the proposed LLSPIN's gates (Right, $>0$ for open gates and $0$ for closed). The values across the x-axis correspond to the feature indices, and the values across the y-axis correspond to the sample indices. Samples are sorted based on their ground truth groups.}%
    \label{fig:lin_gate_com}
\end{figure}

For the motivating example in section \ref{sec:mot_exp}, we used a training set with only $10$ samples. As shown in Fig. \ref{fig:lin_gate_com}, LLSPIN correctly identifies the sample-specific features.

\begin{figure}[htb!]%
    \centering
    \subfloat[\centering ]{{\includegraphics[width=11.2cm]{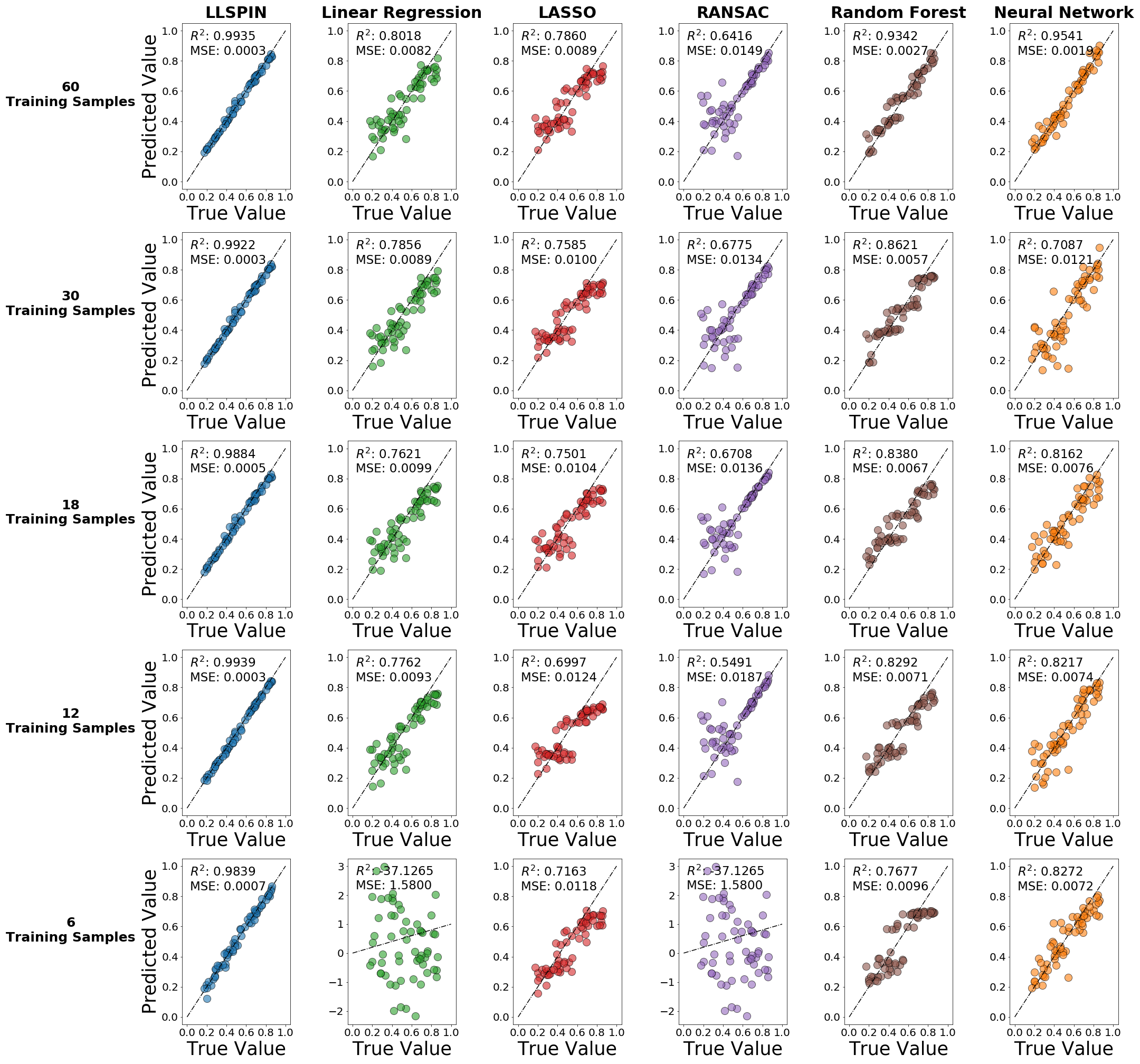}\label{fig:lin_per_comp_others} }}%
    \quad
    \subfloat[\centering  ]{{\includegraphics[width=3.2cm]{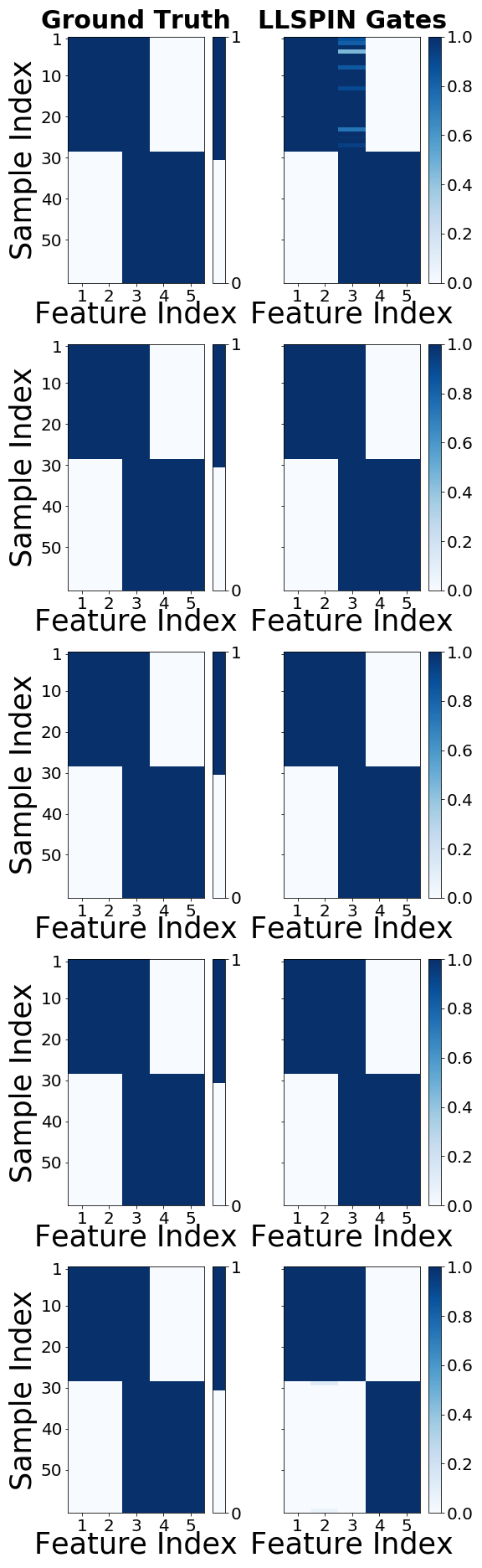}\label{fig:lin_gate_com_others} }}%
    \caption{(a) Evaluation of the performance for different training sizes. Each row indicates a different number of training samples, from top to bottom: 60, 30, 18, 12, and 6, respectively. The x-axis represents the true target value $y$, and the y-axis presents the predicted value $\hat{y}$ for each model (as indicated in the subtitles of the columns). Points on the diagonal line indicate correct predictions. R-square and Mean Squared Error (MSE) are reported for each model. \\(b) Heat-map comparison between the ground truth informative features for each sample (Left, $1$ for truly informative and $0$ for not) and the identified features by LLSPIN's local gates (Right, $>0$ for open gates and $0$ for closed gates). The values across the x-axis correspond to the feature indices, and the values across the y-axis correspond to the sample indices. Samples are sorted based on their group assignments (see model description in Eq. \ref{eq:linear_regression_syn} in the main text).}%
    \label{fig:linear_comp_more_others}%
\end{figure}

Next, we extend our evaluation to other numbers of training samples ($60$,$30$,$18$,$12$,$6$).
Fig. \ref{fig:lin_per_comp_others} demonstrates that LLSPIN consistently outperforms other methods. Fig. \ref{fig:lin_gate_com_others} reveals that in each case, LLSPIN correctly uncovers the corresponding interpretable features for each sample in the gate matrices compared to the ground truth, except when the training is limited with just $6$ training samples where LLSPIN misses one feature for the second sample group. This simulation demonstrates the effectiveness of LLSPIN on low sample size (LSS) datasets. Details of the data model, data split, and hyper-parameter tuning are in the Appendix section \ref{sec:linear_details}.

\subsection{Additional Experiment Involving Linear Synthetic Dataset with Unequal Regression Coefficients}
\label{sec:add_lin}
To further demonstrate our approach's applicability in a more challenging setting, we modified the linear synthetic dataset (Eq. \ref{eq:linear_regression_syn} in the main text). Specifically, the coefficient of $\myvec{x}_3$ in group $2$ is set to $0.5$ instead of $-0.5$, while the coefficient of $\myvec{x}_3$ in group $1$ remains $-0.5$, as shown below:

\begin{align}
    \textnormal{y} &= 
    \begin{cases}
        -2\myvec{x}_1 + \myvec{x}_2 - 0.5\myvec{x}_3 ,&\text{if in group 1,}\\ 
        0.5\myvec{x}_3 + \myvec{x}_4 - 2\myvec{x}_5 ,& \text{otherwise.}\\
    \end{cases}
    \label{eq:linear_regression_syn_uneql}
\end{align}

We note that to learn this more complex regression function (due to the unequal feature coefficients), we had to change the prediction model to a fully connected deep network with nonlinear activation (since the regression function is no longer linear). We applied our nonlinear model (LSPIN) on this example and obtained a $100\%$ true positive rate and $0\%$ false discovery rate in terms of discovering the correct features on the test set. The mean square error on the test is $0.000199$.

\subsection{Nonlinear Synthetic Datasets}

\begin{figure}[htb!]%
    \centering
    \subfloat[\centering \textbf{E1}]{{\includegraphics[width=0.32\textwidth]{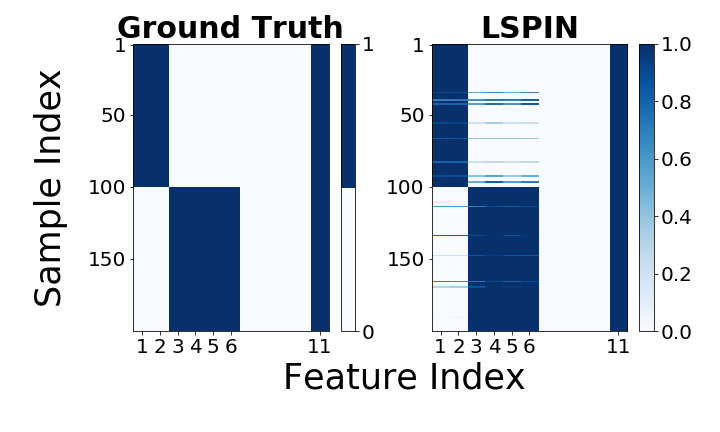} }}%
    \quad
    \subfloat[\centering \textbf{E2}]{{\includegraphics[width=0.32\textwidth]{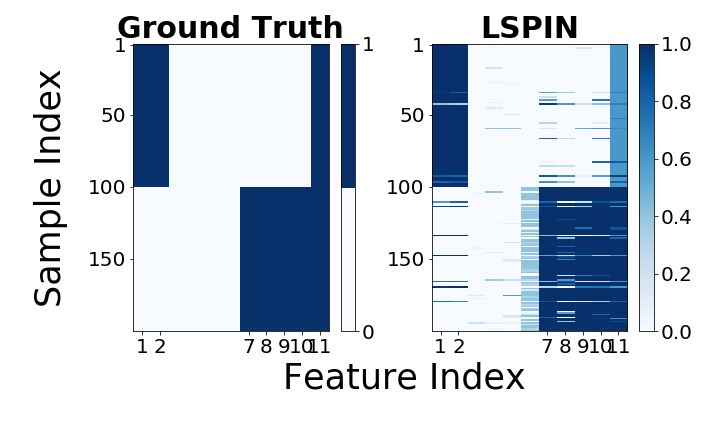} }}%
    \quad
    \subfloat[\centering \textbf{E3}]{{\includegraphics[width=0.32\textwidth]{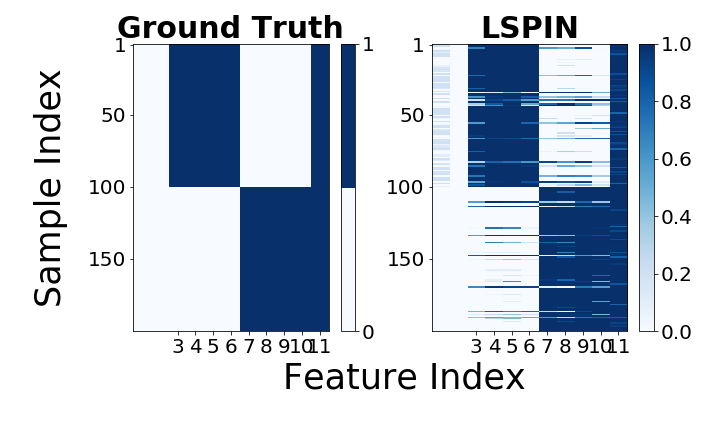} }}%
    \quad
    \subfloat[\centering \textbf{E4}]{{\includegraphics[width=0.32\textwidth]{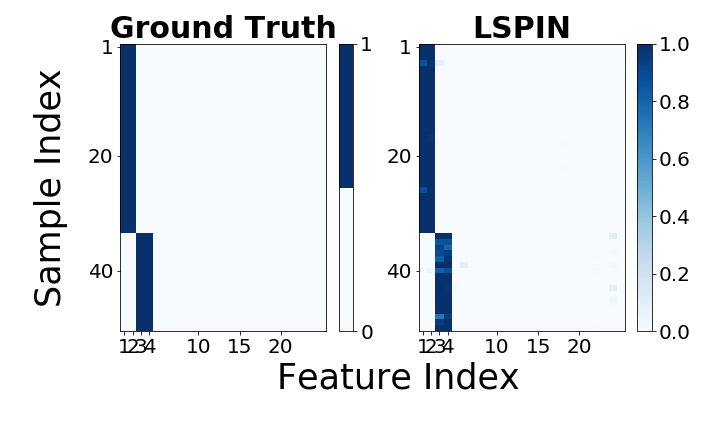} }}%
    
    \subfloat[\centering \textbf{E5}]{{\includegraphics[width=0.32\textwidth]{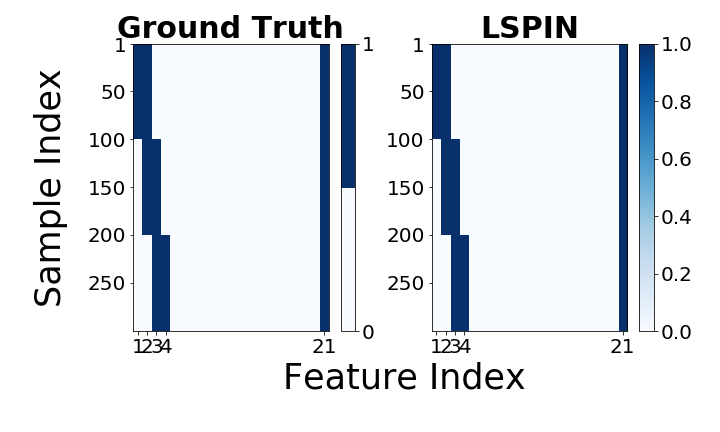} }}%
    
    \caption{Informative feature identification using the nonlinear synthetic datasets. We present heat maps comparing the ground truth informative features (the left panel in each subfigure) and identified features by LSPIN based on the $5$ synthetic datasets (see description in Appendix section \ref{sec:nonl_details}). For more convenient visualization, we only present the first $25$ features for \textbf{E4}.}
    \label{fig:nonl_combo}
\end{figure}

In the following subsection, we first demonstrate LSPIN's interpretability capabilities. We examine the sparsification patterns of LSPIN on the test sets for the 5 nonlinear synthetic examples (see Section \ref{sec:synt} and the Appendix section \ref{sec:nonl_details} for details), as shown in Fig. \ref{fig:nonl_combo} where LSPIN correctly identifies the informative features in most of the examples.

Next, we aim to demonstrate the applicability of our approach in a scenario with overlapping features; we experimented with a modified version of \textbf{E1} (see Eq. \ref{eq:exp1} in the Appendix section \ref{sec:nonl_details}) by changing the model to consist of two overlapping features. Specifically, the $\textnormal{Logit}(\myvec{x})$ in the new function we evaluate becomes: 

\begin{align}
    \textnormal{Logit}(\myvec{x}) &=
    \begin{cases}
        e^{(\myvec{x}_1 \times \myvec{x}_3)},&\text{if } \myvec{x}_{11}<0,\\ 
        e^{(\sum_{i=3}^{6} \myvec{x}_i^2 - 4)},& \text{otherwise}\\
    \end{cases}  \label{eq:exp1_extend}
\end{align}

We note that in this example, both $x_{11}$ and $x_3$ are overlapping. $x_3$ has an unequal factor, and the nonlinear function is not the same in both groups. We applied LSPIN and obtained a true positive rate of $94.68\%$, a false discovery rate of $12.58\%$ on the test set, and test accuracy of $93.00\%$.

For this modified version of \textbf{E1}, we generate data of $6000$ samples by $11$ features with $90\%$ as the training set, $5\%$ as the validation set, and $5\%$ as the test set. In this example, we set the architecture of LSPIN to $5$ hidden layers with $100$ neurons in each layer in the \textit{prediction} network. The number of hidden layers in the \textit{gating} network is $3$ with $100$ neurons in each layer. We set the batch size to $1000$ for training. $\lambda$ is set to $0.15$, the learning rate is set to $0.08$, and the number of epochs is set to $5000$.

    

\subsection{Diversity Evaluation Using the Jaccard Index}\label{sec:diverse}
We expect an excellent interpretability model to identify different sets of variables as driving factors for explaining distinct classes. To evaluate the diversity of all models, we propose using the following Jaccard-based metric. 

First, we compute for each method the per class median set of active features (as indicated by the gates for our method). We denote this set for class $c_i,i=1,...,M$, as ${\cal{S}}_{c_i}$. Then, for each pair of classes we compute intersection using the Jaccard index, namely
$$J({\cal{S}}_{c_i},{\cal{S}}_{c_j})=\frac{|{\cal{S}}_{c_i}\cap {\cal{S}}_{c_j}|}{|{\cal{S}}_{c_i}\cup {\cal{S}}_{c_j}|},i\neq j, $$
then we sum over all possible pairs and normalize by the size of this set, and scale it to $[0,100]$, specifically
$$Diversity=100(1-\sum_{i\neq j}\frac{J({\cal{S}}_{c_i},{\cal{S}}_{c_j})}{M(M-1)/2)}).$$ This quantifies what is the portion of non overlapping features between distinct classes.
\subsection{Extended results for the MNIST Experiment}

\begin{figure}[htb!]
    \centering
    \includegraphics[width=0.48\textwidth]{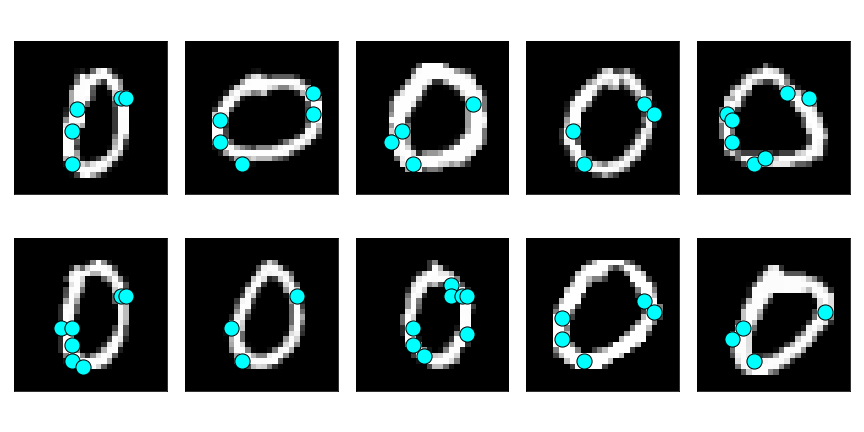} 
    \includegraphics[width=0.48\textwidth]{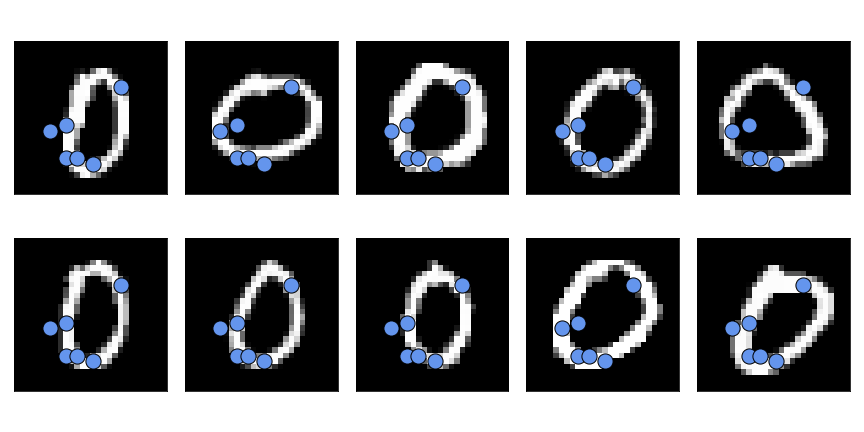} 
    \vskip -0.2 in
    \includegraphics[width=0.48\textwidth]{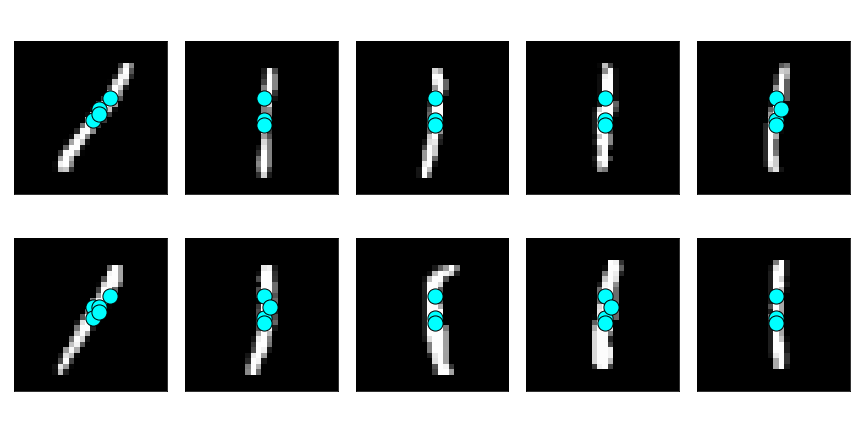} 
    \includegraphics[width=0.48\textwidth]{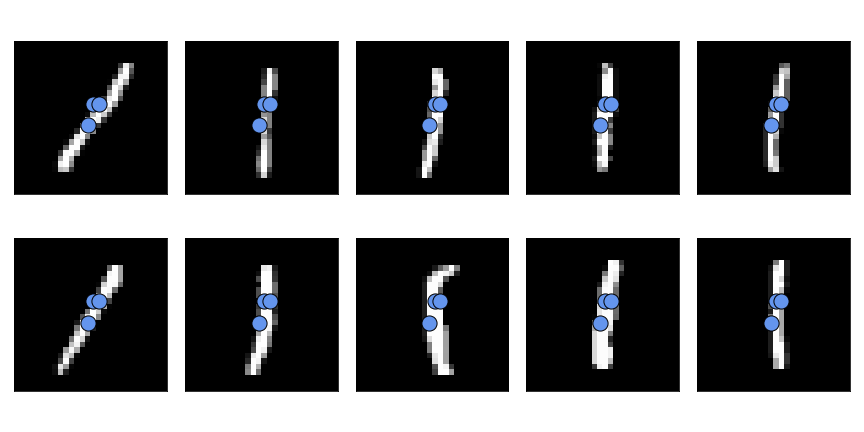}
        \vskip -0.2 in
        \includegraphics[width=0.48\textwidth]{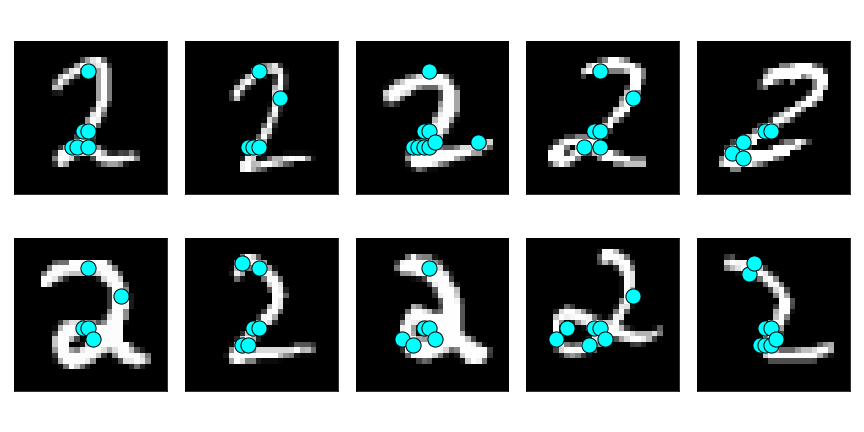} 
    \includegraphics[width=0.48\textwidth]{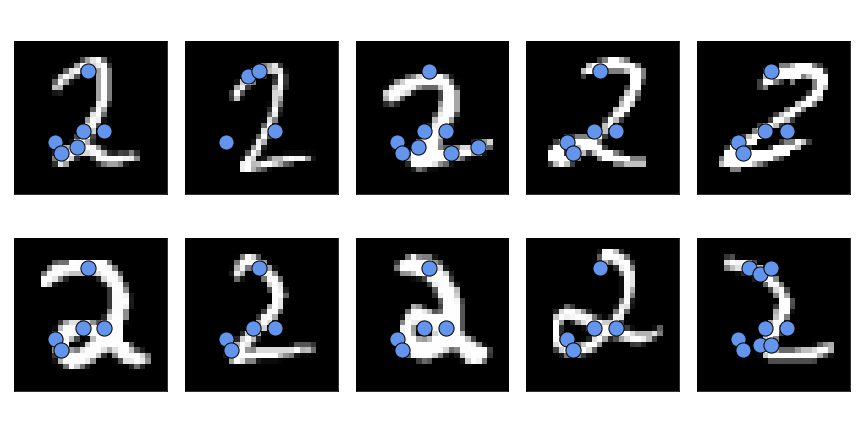} 
    \vskip -0.2 in
    \includegraphics[width=0.48\textwidth]{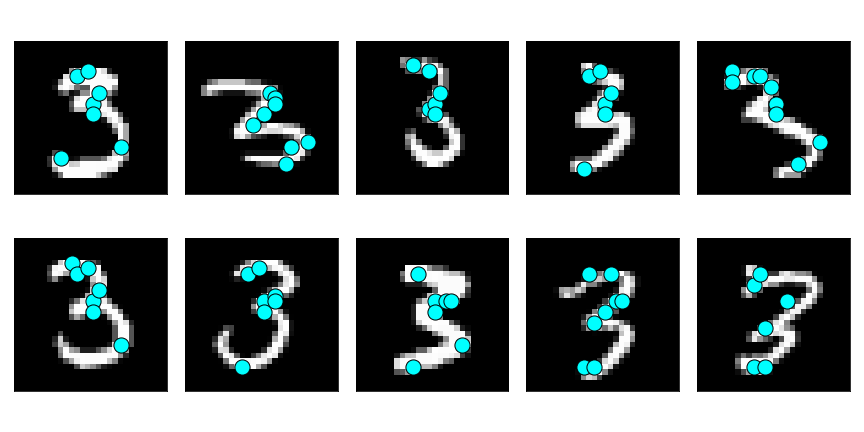} 
    \includegraphics[width=0.48\textwidth]{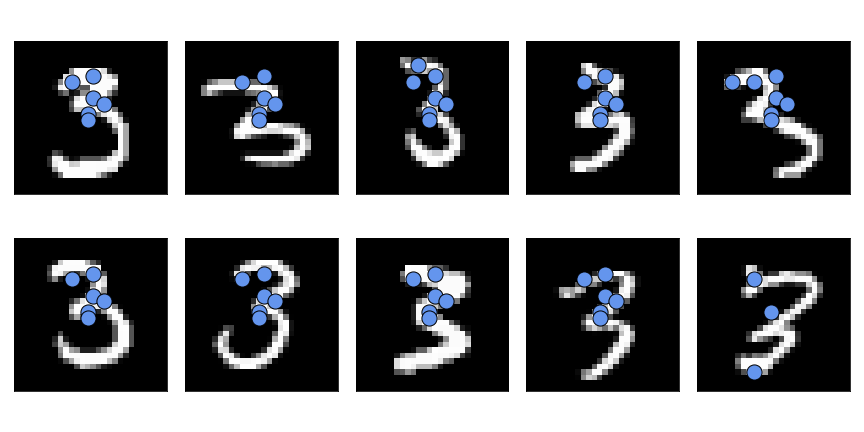} 
\caption{Random representative samples of $0-3$ from the MNIST dataset. For each example, we overlay the image with active gates, here color dots indicate locations of non zero gates. Left columns (cyan dots) represents the selected pixels with LLSPIN using $\lambda_2=0$. Right columns (darker blue dots) represents the selected pixels with LLSPIN using $\lambda_2=0.1$. Notice that when we increase $\lambda_2$ the selection becomes more stable across samples within a class. }
\label{fig:mnist_additional_1}
\end{figure}

\begin{figure}[htb!]
    \centering
  \includegraphics[width=0.48\textwidth]{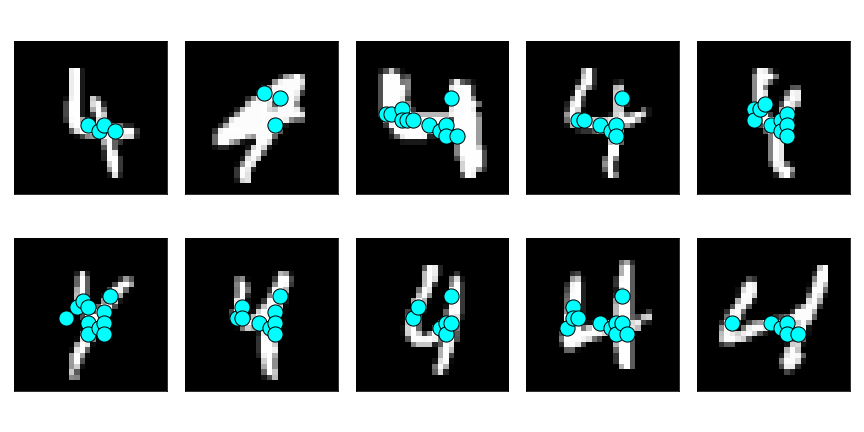} 
    \includegraphics[width=0.48\textwidth]{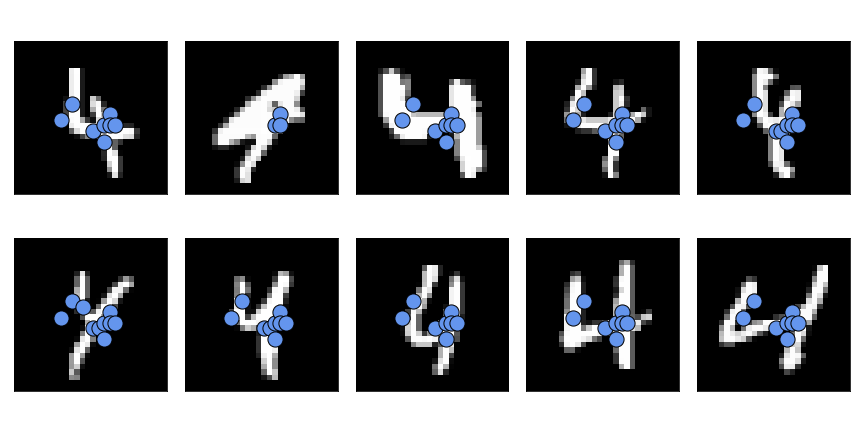} 
    \vskip -0.2 in
    \includegraphics[width=0.48\textwidth]{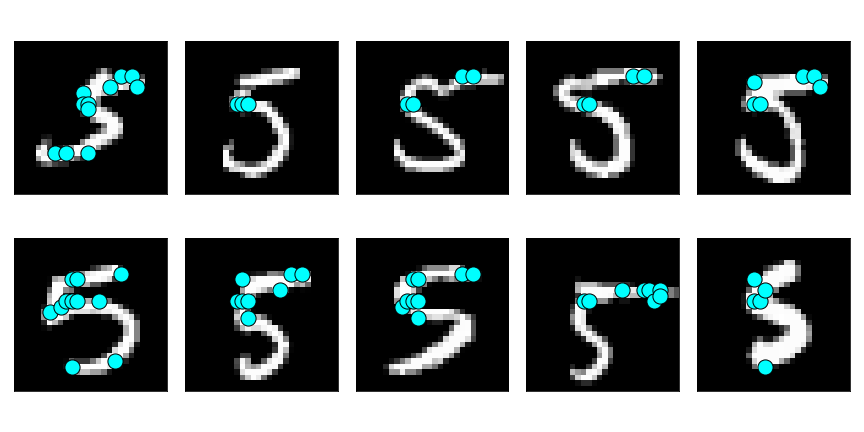} 
    \includegraphics[width=0.48\textwidth]{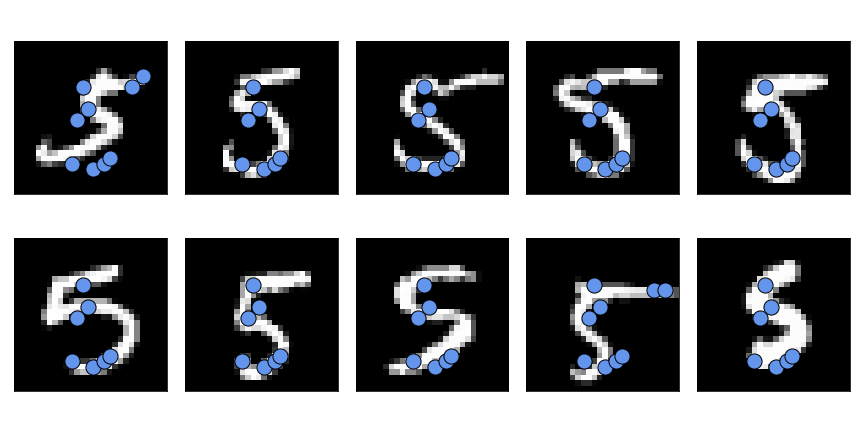}
        \vskip -0.2 in
        \includegraphics[width=0.48\textwidth]{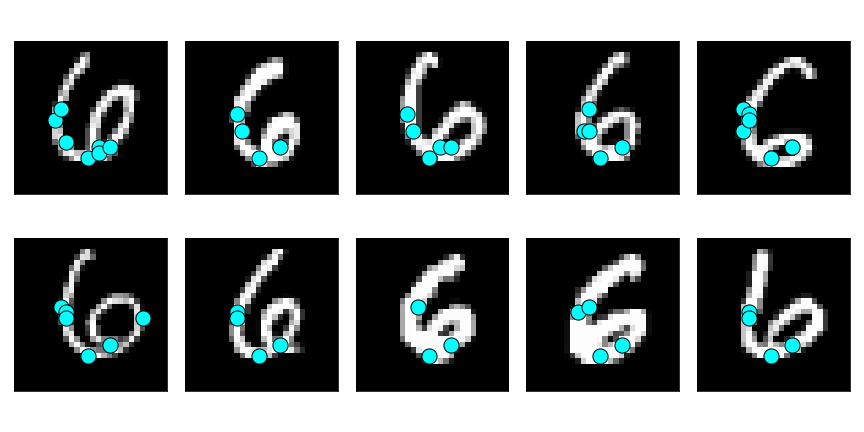} 
    \includegraphics[width=0.48\textwidth]{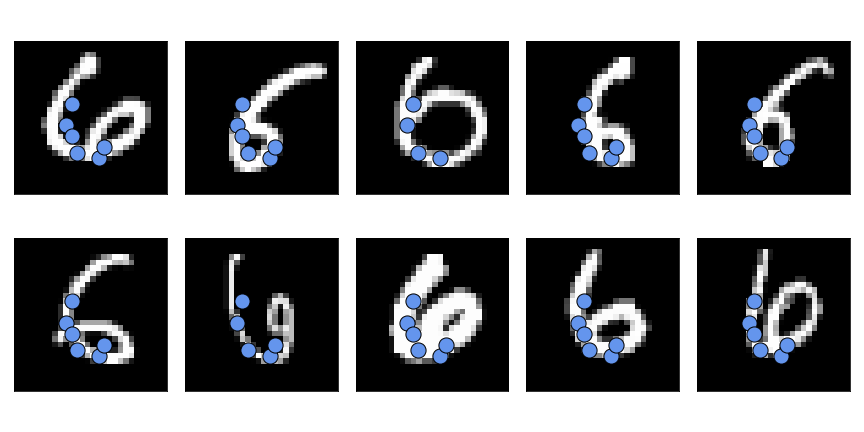} 
    \vskip -0.2 in
    \includegraphics[width=0.48\textwidth]{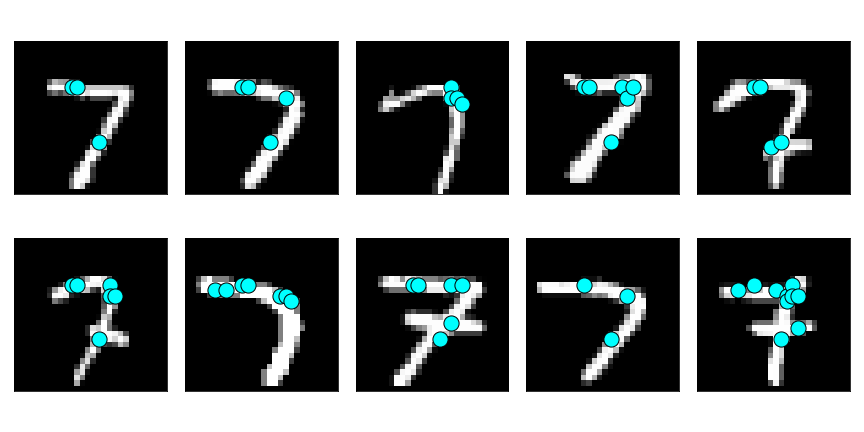} 
    \includegraphics[width=0.48\textwidth]{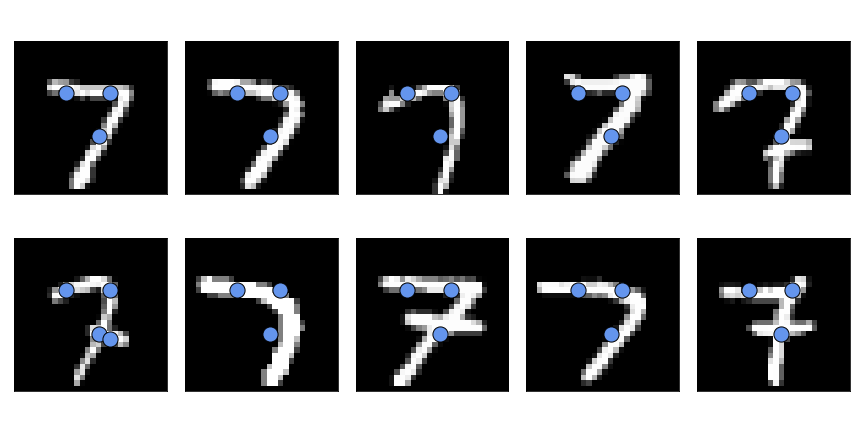} 
\caption{Random representative samples of $4-7$ from the MNIST dataset. We overlay the image with active gates locations for each example, indicated here as color dots. Left columns (cyan dots) represents the selected pixels with LLSPIN using $\lambda_2=0$. Right columns (darker blue dots) represents the selected pixels with LLSPIN using $\lambda_2=0.1$. Notice that when we increase $\lambda_2$ the selection becomes more stable across samples within a class.}
\label{fig:mnist_additional_2}
\end{figure}

\begin{figure}[htb!]
    \centering
  \includegraphics[width=0.48\textwidth]{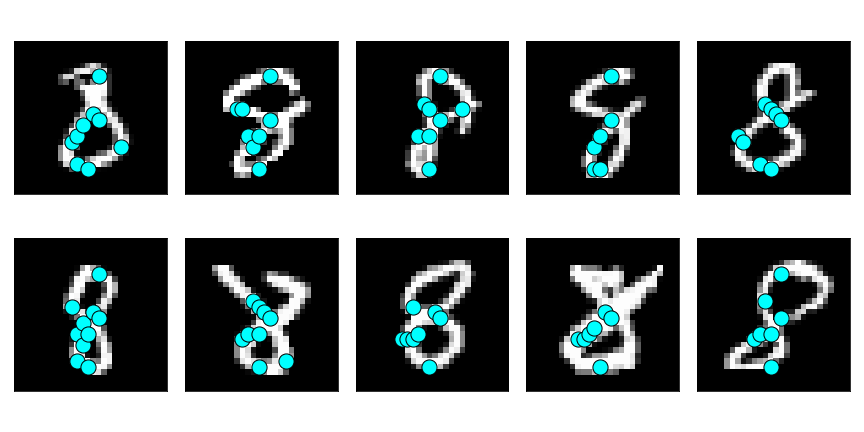} 
    \includegraphics[width=0.48\textwidth]{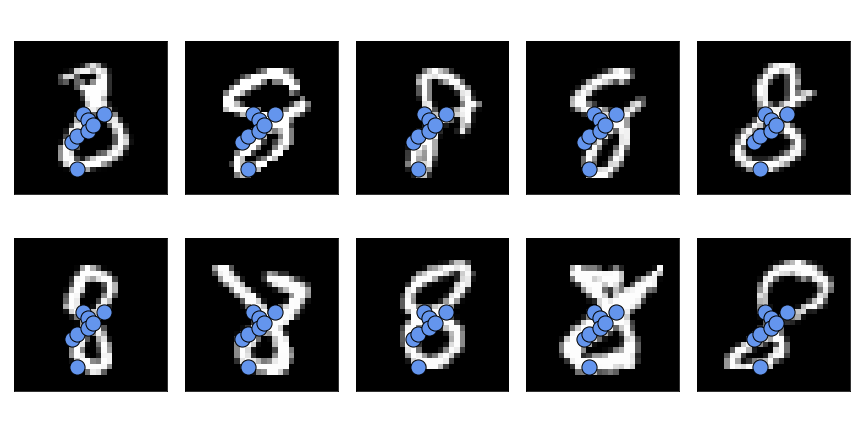} 
    \vskip -0.2 in
    \includegraphics[width=0.48\textwidth]{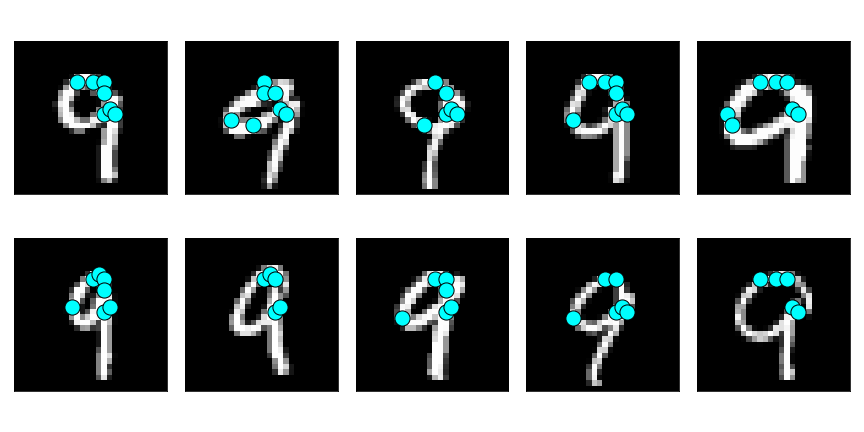} 
    \includegraphics[width=0.48\textwidth]{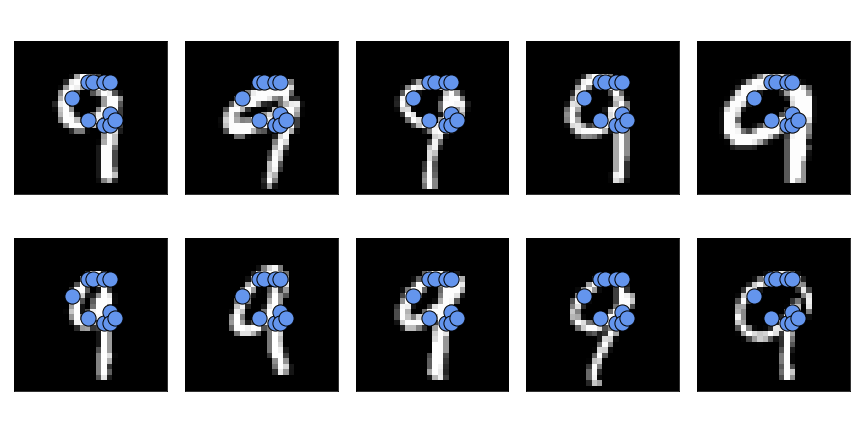}

\caption{Random representative samples of $8-9$ from the MNIST dataset. We overlay the image with active gates locations for each example, indicated here as color dots. Left columns (cyan dots) represents the selected pixels with LLSPIN using $\lambda_2=0$. Right columns (darker blue dots) represents the selected pixels with LLSPIN using $\lambda_2=0.1$. Notice that when we increase $\lambda_2$ the selection becomes more stable across samples within a class.}
\label{fig:mnist_additional_3}
\end{figure}

\begin{figure}[htb!]
    \centering
  \includegraphics[width=0.48\textwidth]{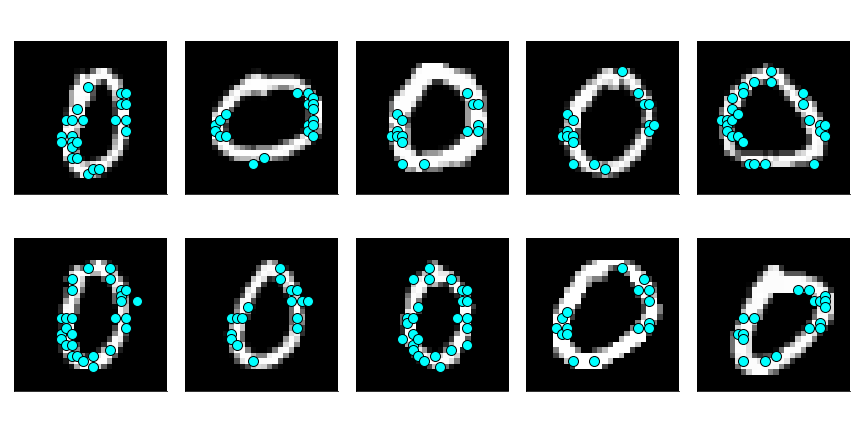} 
    \includegraphics[width=0.48\textwidth]{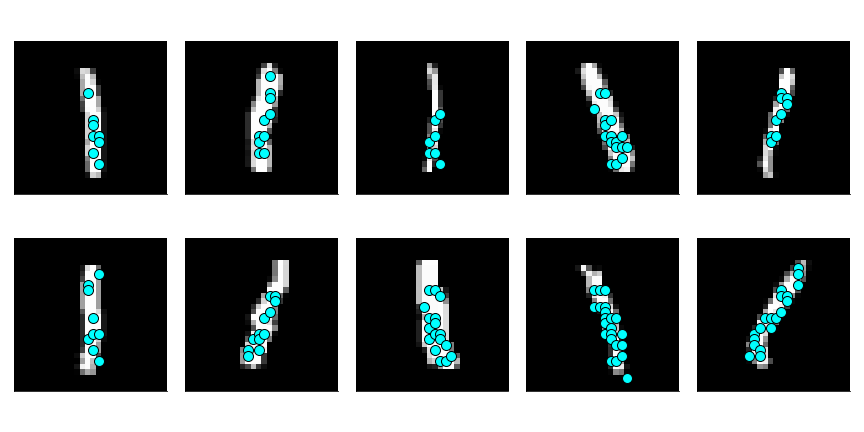} 

\caption{Random representative samples of $0-1$ from the MNIST dataset. We overlay the image with active gates locations for each example, indicated here as color dots. Here, we tune the model to select a median of $24$ informative features per image.}
\label{fig:mnist_additional_4}
\end{figure}

Extending from the MNIST experiment demonstrated in Section \ref{sec:mnist_exp}, here we present additional randomly selected images for the different MNIST classes and superimpose the images using gates with non zero values in Figs. \ref{fig:mnist_additional_1},\ref{fig:mnist_additional_2}, and \ref{fig:mnist_additional_3}. In these example we highlight the effect of adding the second regularization term, controlled by $\lambda_2$. Finally, in Fig. \ref{fig:mnist_additional_4} we present the selected pixels when we tune our model to select more features per sample.

\subsection{Experiment with Low Sample Size MNIST Dataset}

To further test the performance of our method on the LSS regime, we evaluate LLSPIN on a low sample size MNIST variant. Specifically, we use 6k samples for training and 10k samples for testing. A fully connected MLP with four layers [784,300,100,10] and a tanh activation leads to a test accuracy of $94.2\%$. Using the linear variant of our method LLSPIN, we reach a test accuracy of $94.8\%$ while using a median of $10$ pixels per image with no activation in the \textit{prediction} network. In comparison, this may seem like a minor improvement and far from state-of-the-art (which requires a convolution layer). However, we argue that because our prediction model is linear, we can easily interpret each prediction since we obtain a small subset of active pixels and their linear coefficient values.

\subsection{Identification of Cell-type Specific Markers}

\begin{figure*}[htb!]
    \centering
    \includegraphics[width=0.8 \textwidth]{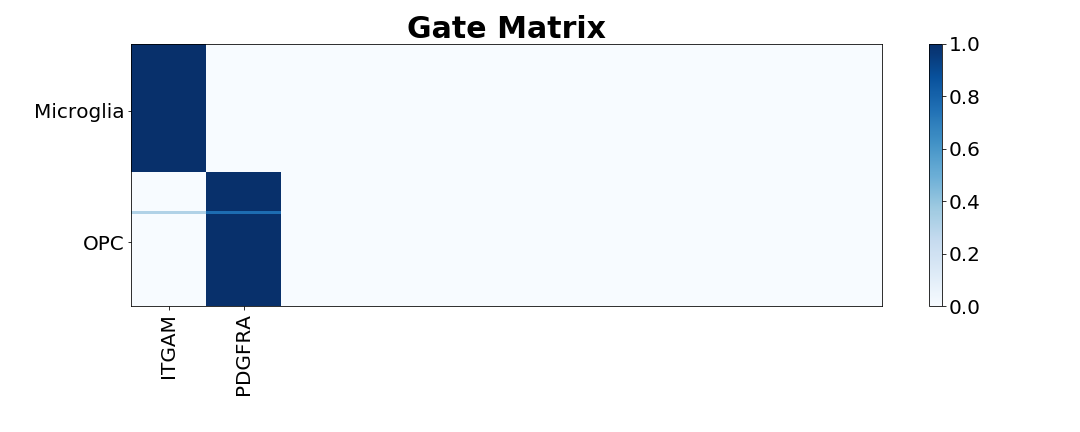}
    \caption{Identification of the cell-type-specific markers by LLSPIN's local gates ($> 0$ for open gates and $0$ for closed gates). The values across the x-axis correspond to the different genes (with only the $2$ marker genes shown), and the values across the y-axis correspond to the different cells (sorted by the cell types: Microglia and Oligodendrocyte Precursor Cells (OPC)). For more convenient visualization, we only present the first $30$ genes.}
    \label{fig:cell_type_markers}
\end{figure*}

In this subsection, we examine LLSPIN's interpretability capabilities on the Single Nucleus RNA-sequencing dataset described in Section \ref{sec:sn_results}. As shown in Fig. \ref{fig:cell_type_markers}, LLSPIN identifies the correct cell type specific marker genes for each cell type group (\textit{ITGAM} gene for Microglia cells and \textit{PDGFRA} for Oligodendrocyte Precursor Cells (OPC)).

\subsection{Sparsity of the Local Gates}

To demonstrate that our model performs local sparsification of input features, we evaluate the statistics of the gate values. Table \ref{tab:gate_convergence} presents the distribution of local gates values in some of the synthetic experiments. Our results demonstrate that most gates converge to $0$ and $1$ with only a few gates converging to values in the range $(0,1)$.

\begin{table}[htb!]
    \centering
    \begin{adjustbox}{max width=1 \textwidth,min height = 0.6 in, valign=l}
    \begin{tabular}{|c|c|c|c|c|c|c|c|}
    \hline
                &  & \multicolumn{3}{c|}{Training} & \multicolumn{3}{c|}{Test}  \\
    \hline
    Experiments & Models & \% of 0s & \% of 1s & \textbf{\% 0-1}  & \% of 0s & \% of 1s & \textbf{\% 0-1} \\
    \hline
    Linear 60 samples & LLSPIN & 40.00 & 57.67 & \textbf{2.33} & 40.00 & 57.33 & \textbf{2.67} \\
    \hline
    Linear 30 samples & LLSPIN & 40.00 & 60.00 & \textbf{0.00} & 40.00 & 60.00 & \textbf{0.00} \\
    \hline
    Linear 18 samples & LLSPIN & 40.00 & 60.00 & \textbf{0.00} & 40.00 & 60.00 & \textbf{0.00} \\
    \hline
    Linear 12 samples & LLSPIN & 40.00 & 60.00 & \textbf{0.00} & 40.00 & 60.00 & \textbf{0.00} \\
    \hline
    Linear 10 samples & LLSPIN & 40.00 & 58.00 & \textbf{2.00} & 40.00 & 59.33 & \textbf{0.67} \\
    \hline
    Linear 6 samples & LLSPIN & 50.00 & 50.00 & \textbf{0.00} & 50.00 & 49.33 & \textbf{0.67} \\
    \hline
    \multirow{2}{*}{Nonlinear \textbf{E1}} & LLSPIN & 91.11 & 8.60 & \textbf{0.29} & 91.14 & 7.27 & \textbf{1.59}\\\cline{2-8}
                               & LSPIN & 62.90 & 36.54 & \textbf{0.56} & 62.27 & 37.00 & \textbf{0.73}\\
   \hline
    \multirow{2}{*}{Nonlinear \textbf{E2}} & LLSPIN & 78.36 & 20.68 & \textbf{0.96} & 78.59 & 20.09 & \textbf{1.32}\\\cline{2-8}
                               & LSPIN & 63.25 & 35.53 & \textbf{1.22} & 62.91 & 35.59 & \textbf{1.50}\\
   \hline
    \multirow{2}{*}{Nonlinear \textbf{E3}} & LLSPIN & 83.67 & 13.69 & \textbf{2.63} & 83.36 & 13.32 & \textbf{3.32}\\\cline{2-8}
                               & LSPIN & 52.04 & 44.90 & \textbf{3.07} & 51.82 & 44.73 & \textbf{3.45} \\
   \hline
    \multirow{2}{*}{Nonlinear \textbf{E4}} & LLSPIN & 95.50 & 3.56 & \textbf{0.94} & 95.68 & 3.28 & \textbf{1.04}\\\cline{2-8}
                               & LSPIN & 95.82 & 3.74 & \textbf{0.44} & 95.72 & 3.68 & \textbf{0.60} \\
   \hline
    \end{tabular}
    \end{adjustbox}
    \caption{Statistics of the local gate values. The percentage of gates between $0$ and $1$ is listed in bold.}
    \label{tab:gate_convergence}
\end{table}

\subsection{Evaluation of the Fairness and Robustness of LLSPIN} \label{sec:shift}
Following the analysis in \cite{ovadia2019can}, we conducted several experiments to evaluate how our proposed method performs under distributional shifts. Specifically, we have compared the accuracy, Negative Log-Likelihood (NLL), and Expected Calibration Error (ECE) of our model (LLSPIN) to those obtained by a fully connected nonlinear MLP. Table \ref{tab:fair&robust_1} details the results for different rotation angles of the test samples from MNIST datasets.

Next, we present the NLL results based on the experiments performed on MNIST, low sample size MNIST, and real-world datasets in Table \ref{tab:fair&robust_2}. Based on the results presented in the table, we conclude that our model leads to better-calibrated uncertainty. This is evident by the lower ECE and NLL values our model obtains compared with others.

\begin{table}[htb!]
  \centering
  
    \begin{tabular}{|c|c|c|c|c|c|c|}
    \hline
    \multicolumn{1}{|c|}{Rotation MNIST} & \multicolumn{3}{c|}{LLSPIN} & \multicolumn{3}{c|}{FULL} \\
    \hline
    \multicolumn{1}{|c|}{Angle} & \multicolumn{1}{c|}{ACC} & \multicolumn{1}{c|}{NLL} & \multicolumn{1}{c|}{ECE} & \multicolumn{1}{c|}{ACC} & \multicolumn{1}{c|}{NLL} & \multicolumn{1}{c|}{ECE} \\
    \hline
    10    & \textbf{96.9}  & \textbf{1.403} & \textbf{0.014} & 96.8  & 1.987 & 0.023 \\
    \hline
    20    & \textbf{92.1}  & \textbf{3.777} & \textbf{0.039} & 90.8  & 6.455 & 0.071 \\
    \hline
    30    & \textbf{77.9}  & \textbf{10.341} & \textbf{0.126} & 75.8  & 19.859 & 0.195 \\
    \hline
    40    & \textbf{59.8}  & \textbf{20.532} & \textbf{0.243} & 56.4  & 43.771 & 0.367 \\
    \hline
    50    & \textbf{40.1}  & \textbf{31.006} & \textbf{0.377} & 38.6  & 71.346 & 0.532 \\
    \hline
    60    & \textbf{27.8}  & \textbf{40.641} & \textbf{0.473} & 26.8  & 96.177 & 0.643 \\
    \hline
    70    & \textbf{19.49} & \textbf{48.614} & \textbf{0.543} & 18.4  & 119.586 & 0.724 \\
    \hline
    80    & 14.24 & \textbf{55.249} & \textbf{0.601} & \textbf{14.6}  & 139.701 & 0.769 \\
    \hline
    90    & 11.7  & \textbf{60.477} & \textbf{0.637} & \textbf{12.3}  & 155.147 & 0.795 \\
    \hline
    \end{tabular}%
    \caption{Comparison of classification accuracy (ACC), Negative Log Likelihood (NLL), Expected Calibration Error (ECE) on rotated MNIST.}
  \label{tab:fair&robust_1}%
\end{table}%

\begin{table}[htb!]
  \centering
    \begin{tabular}{|c|c|c|c|}
    \hline
          &  LLSPIN  & FULL  & INVASE \\
    \hline
    MNIST & \textbf{0.826} & 1.043 & 2.008 \\
    \hline
    low sample size MNIST & \textbf{1.985} & 3.311 & 4.675 \\
    \hline
    TOX-171 & \textbf{0.760±0.411} & 3.968±0.500 & 2.593±0.799 \\
    \hline
    RELATHE & \textbf{0.879±0.189} & 1.858±0.047 & 1.098±0.123 \\
    \hline
    BASEHOCK & \textbf{0.367±0.057} & 1.813±0.040 & 0.668±0.103 \\
    \hline
    COLON & \textbf{0.884±0.415} & 1.071±0.394 & 1.318±0.653 \\
    \hline
    PBMC  & \textbf{1.012±0.046} & 4.062±0.059 & 1.441±0.048 \\
    \hline
    PCMAC & \textbf{0.816±0.118} & 1.873±0.049 & 0.905±0.050 \\
    \hline
    \end{tabular}%
     \caption{Comparison of Negative Log-Likelihood (NLL) on MNIST, low sample size MNIST, and the real-world datasets.}
  \label{tab:fair&robust_2}%
\end{table}%

\subsection{Discussion on the Number of Selected Features}
\label{sec:large_features}

When applied to high dimensional data, the model may select a large set of features across some samples. This phenomenon suggests that the model is overfitting. Based on our experience, this happens for a small set of observations in the dataset (both in the train and test). For example, in the PBMC data for $8.6\%$ of the train samples, the model selects more than $10\%$ of the features (genes). This information could indicate that the model may be overfitting on a small subset of samples. Therefore, we believe that the local gates could provide additional information for practitioners to help interpret the sample-specific predictions made by the model instead of just relying on the average accuracy the model obtains on the test set. We did not observe this overfitting phenomenon in some of the low-dimensional datasets evaluated in the paper. For example, on MNIST, the model selects a median of $8$ features, a union of $165$ features. Furthermore, we provide Table \ref{tab:mnist_stat} to demonstrate the statistics of the selected features on MNIST.

\begin{table}[htb!]
    \centering
    \begin{tabular}{|c|c|}
    \hline
    Number of selected features & Percent of samples \\
    \hline
    $<5$ & $5.90\%$ \\
    \hline
    $5 - 9$ & $79.36\%$ \\
    \hline
    $10 - 14$ & $14.50\%$ \\
    \hline
    $15 - 19$ & $0.24\%$ \\
    \hline
    $20+$ & $0.00\%$\\
    \hline
    \end{tabular}
    \caption{Statistics of the selected features on MNIST}
    \label{tab:mnist_stat}
\end{table}

Another example is the SEER cancer dataset; when the model selects a median of 3 features (leftmost point of Fig. \ref{fig:cox_result} (Left)), the union number of selected features is $9$. In Table \ref{tab:seer_stat}, we present the statistics of the features chosen for this example.

In these two examples, the model can also reduce the total number of selected features (since the union of features used by the model is relatively small). 

\begin{table}[htb!]
    \centering
    \begin{tabular}{|c|c|}
     \hline
    Number of selected features & Percent of samples \\
     \hline
     $<3$ & $95.25\%$\\
     \hline
     $3 -5$ & $4.73\%$ \\
     \hline
     $6 -7 $ & $0.02\%$ \\
     \hline
     $8+$ & $0.00\%$ \\
     \hline
    \end{tabular}
    \caption{Statistics of the selected features on the SEER dataset.}
    \label{tab:seer_stat}
\end{table}

\subsection{Time Benchmark Results}
\label{sec:benchmark_time}

\begin{figure}[htb!]%
    \centering
    \subfloat[\centering ]{{\includegraphics[width=7cm]{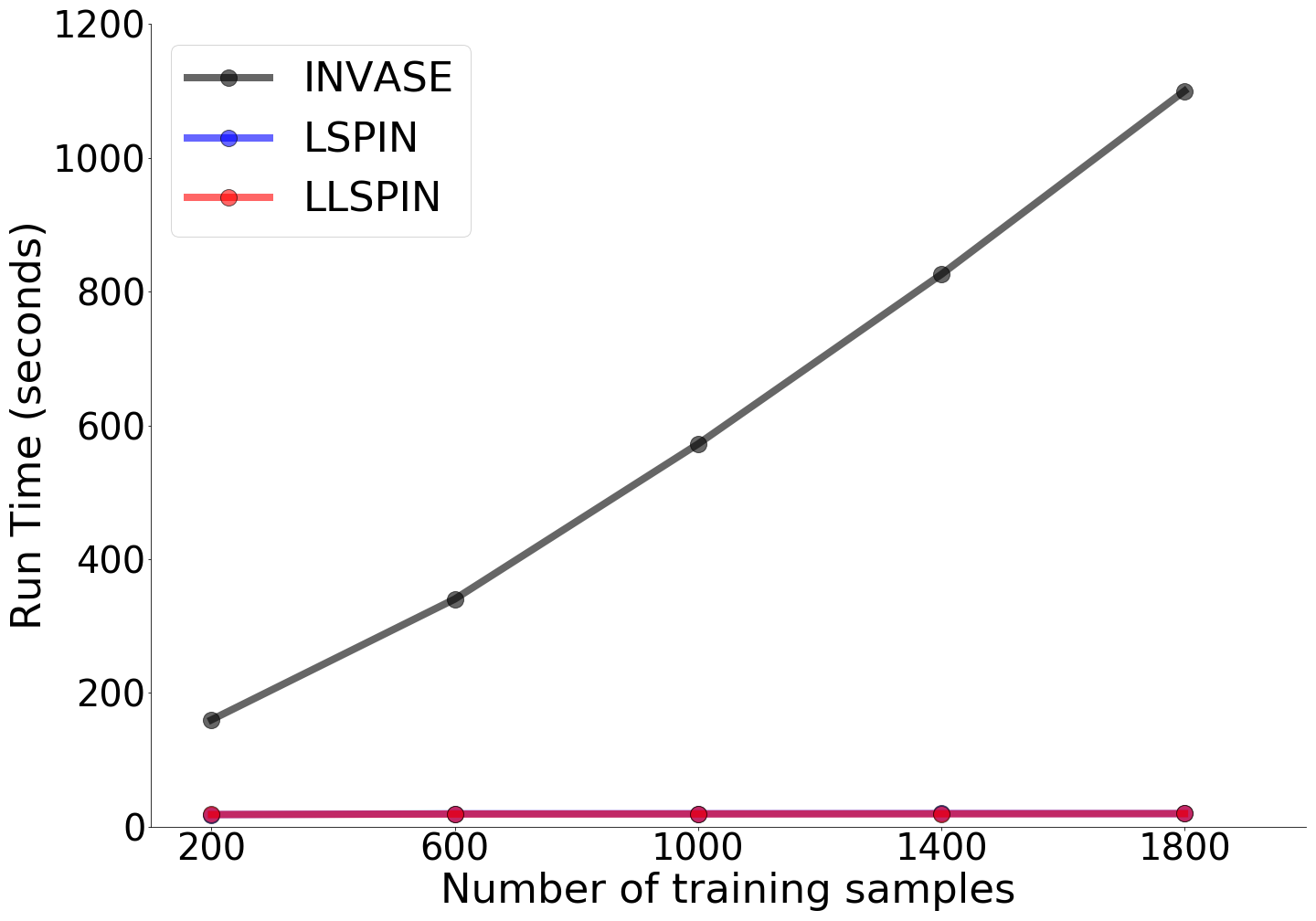}\label{fig:timebench2} }}%
    \quad
    \subfloat[\centering  ]{{\includegraphics[width=7cm]{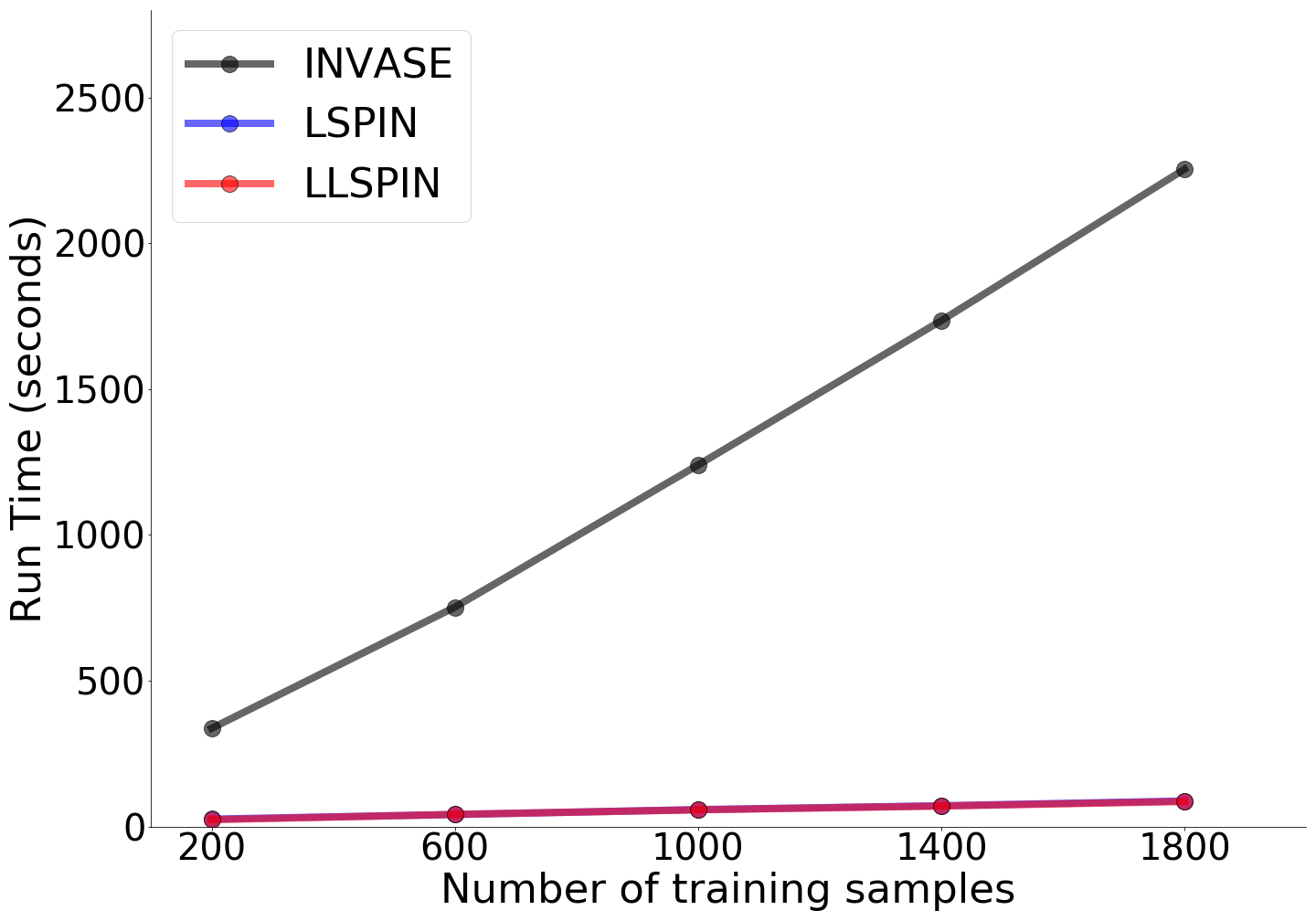}\label{fig:timebench} }}%
    \caption{Time Benchmark between LLSPIN/LSPIN and INVASE on 2 datasets}%
    \label{fig:time_benchmarking}%
\end{figure}

To demonstrate the computational efficiency of our models, we first compare the training running time between LLSPIN/LSPIN and INVASE on the nonlinear synthetic example \textbf{E1} (Eq. \ref{eq:exp1} in Appendix section \ref{sec:nonl_details}) where we vary the number of training samples ($200,600,1000,1400,1800$) as shown in Fig. \ref{fig:timebench2}. We can see that the running time of INVASE increases rapidly with more training samples, whereas our models remain scalable. 

To benchmark the running time when there are more features, we use the dataset from \textbf{E2} (Eq. \ref{eq:exp2} in Appendix section \ref{sec:nonl_details}) and generate additional noisy features ($2000$ features in total including the informative $11$ features) by sampling values from $\mathcal{N}(0,1)$. We show the comparison in Fig. \ref{fig:timebench}. In this high dimensional regime, LLSPIN/LSPIN remain scalable compared with INVASE.

We design LLSPIN/LSPIN/INVASE using $2$ hidden layers with $200$ neurons each to perform a fair comparison. We set $2$ hidden layers with $100$ neurons each for the gating network of LLSPIN/LSPIN and the selector network of INVASE. For all the three models, $\lambda$ is set to 1, the batch size is set to full batch training, and epochs is set to $3000$. We set the learning rate for INVASE to $0.0001$ (Adam optimizer) and $0.1$ for SGD optimizer of LLSPIN/LSPIN.

\renewcommand\thefigure{\thesection.\arabic{figure}} 
\setcounter{figure}{0}  
\renewcommand\thetable{\thesection.\arabic{table}} 
\setcounter{table}{0}  

\section{Reproducibility and Additional Details}
\label{sec:reproduce}

In the following subsections, we provide additional experimental details required for the reproduction of the experiments provided in the main text. The CPU model used for the experiments is Intel(R) Xeon(R) Gold 6150 CPU @ 2.70GHz (72 cores total). GPU model is NVIDIA GeForce RTX 2080 Ti. The operating system is Ubuntu 20.04.2 LTS. The memory storage is 1 TB in total. The software dependencies are specified in the associated codes.  

We apply batch normalization to the prediction network of LLSPIN/LSPIN and STG models throughout the experiments, except for the MNIST example and the survival analysis. We note that the application of batch normalization to the MNIST data did not improve performance. In the survival analysis, the performance was satisfactory without the application of batch normalization.

For LLSPIN/LSPIN/STG models, the network weights are initialized by drawing $\mathcal{N}(0,s)$ and bias terms are set to $0$. For LLSPIN/LSPIN, we set $s$ to be $0.1$ for the MNIST example, time benchmark experiments, and the synthetic datasets experiments (Except for LSPIN on \textbf{E1} and LLSPIN/LSPIN on \textbf{E3} in Section \ref{sec:synt} where $s$ is set to $0.05$ for better convergence). We set $s = \frac{1}{\sqrt{D}}$ ($D$ is the input dimensionality) following Xaiver initialization \cite{glorot2010understanding} for the real-world examples, and $s = 0.05$ for the cox survival analysis.  For marker gene identification, $s$ in the prediction network is $0.1$, and $s$ in the gating network is set to $0.001$ which we found to be helpful for stabilizing the training. For STG models, $s$ is set to $0.1$ for the real-world datasets and $0.05$ for the cox survival analysis.
Unless expressly noted, we use tanh as the hidden layer activation function for both the gating network of LLSPIN/LSPIN and the nonlinear prediction network of LSPIN/STG. For other neural network-based methods, we use their default activation functions. 
   
Across the experiments, we only enabled the second regularization term in the MNIST example and the marker gene identification example.   

For the experiments, standard metrics in supervised learning, including classification accuracy, $R^2$ and mean squared error and concordance index, are adopted to evaluate the performance of different models in classification, regression, and survival analysis tasks.

\subsection{Algorithms}

\begin{algorithm}[t!]
\vskip -0.05 in
   \caption{ Locally SParse Interpretable Networks (LSPIN) Pseudo-code}
   \label{alg:pseudocode}
\begin{algorithmic}
\STATE {\bfseries Training:}
   \STATE {\bfseries Input:} observations $\{\myvec{x}^{(i)},y^{(i)}\}^N_{i=1}$, regularization parameter $\lambda$, number of epochs $T$, batch size $B$,
   learning rate $\gamma$.
   \STATE {\bfseries Output:} Gating network $\myvec{\Psi}_\myvec{\Omega}$ and prediction model $\myvec{f}_{\myvec{\theta}}$
   
   \STATE Initialize the weights $\myvec{\Omega}$ of gating network $\myvec{\Psi}$
   
   \FOR{$t=1$ {\bfseries to} $T$}
     \FOR{each size B batch}
       \FOR{$i=1$ {\bfseries to} $B$}
       \STATE Compute ${\myvec{\mu}^{(i)}}=\myvec{\psi}(\myvec{x}^{(i)}|\myvec{\Omega})$
       \STATE Sample ${\mathbf{\epsilon}}^{(i)}$ from $\mathcal{N}(0,\myvec{I}\sigma^2)$
       \STATE Compute local stochastic gates: 
        \STATE \quad $\mathbf{z}^{(i)} = \max(0,\min(1,0.5+ \myvec{\mu}^{(i)} + \mathbf{\epsilon}^{(i)}))$
        \ENDFOR
       \STATE Compute the loss: \\ $\myvec{\tilde{L}} = \frac{1}{B} \sum_{i=1}^{B}( \|f_{\myvec{\theta}}(\myvec{x}^{(i)}\odot\mathbf{z}^{(i)})-y^{(i)} \|_2 +  \myvec{R(\mathbf{z}^{(i)})})$ (where $\myvec{R(\mathbf{z}^{(i)})}$ is defined in Eq. \ref{eq:reg})
       \STATE Update $\myvec{\theta}= \myvec{\theta} -  \gamma \nabla_{\myvec{\theta}} \myvec{\tilde{L}}$,\quad $\myvec{\Omega} = \myvec{\Omega} - \gamma \nabla_{\myvec{\Omega}}  \myvec{\tilde{L}}$
      \ENDFOR
    \ENDFOR
   
\STATE

\STATE {\bfseries Inference:}
    \STATE {\bfseries Input:} observations $\{\myvec{x}^{(i)}\}^M_{i=1}$ with $\myvec{x}^{(i)}\in \mathbb{R}^D$, Trained gating network $\myvec{\Psi}_\myvec{\Omega}$, and prediction model $f_{\myvec{\theta}}$
   \STATE {\bfseries Output:} Local gates: 
   \STATE \quad \quad \quad \quad $\{\myvec{z}^{(i)} = \max(0,\min(1,0.5+ \myvec{\psi}(\myvec{x}^{(i)}|\myvec{\Omega}) ))\}^M_{i=1}$
 \newline \hspace*{3.5em}
 Predictions: $\{\tilde{y}^{(i)} = f_{\myvec{\theta}}(\myvec{x}^{(i)}\odot\myvec{z}^{(i)})\}^M_{i=1}$
\vskip -0.2 in
\end{algorithmic}

\end{algorithm}

\subsection{Regularization Term}

The leading term in our regularizer is expressed by : 
\begin{align*}
    \mathbb{E}_Z \norm{\myvec{Z}}_0 &= \sum_{d=1}^D \mathbb{P}[z_d > 0] = \sum_{d=1}^D \mathbb{P}[\mu_d + \sigma \epsilon_d +0.5> 0] \\
    &= \sum_{d=1}^D \{ 1 - \mathbb{P}[\mu_d + \sigma \epsilon_d +0.5 \le 0] \} \\
    &= \sum_{d=1}^D \{ 1 - \Phi(\frac{-\mu_d-0.5}{\sigma}) \} \\
    &= \sum_{d=1}^D \Phi\left (\frac{\mu_d+0.5}{\sigma} \right) \\
    &= \sum^D_{d=1}\left(\frac{1}{2} - \frac{1}{2}\erf\left(-\frac{\mu_d+0.5}{\sqrt{2}\sigma}\right) \right)
\end{align*}

\begin{figure}[htb!]
\begin{center} 
\includegraphics[width=0.5\textwidth]{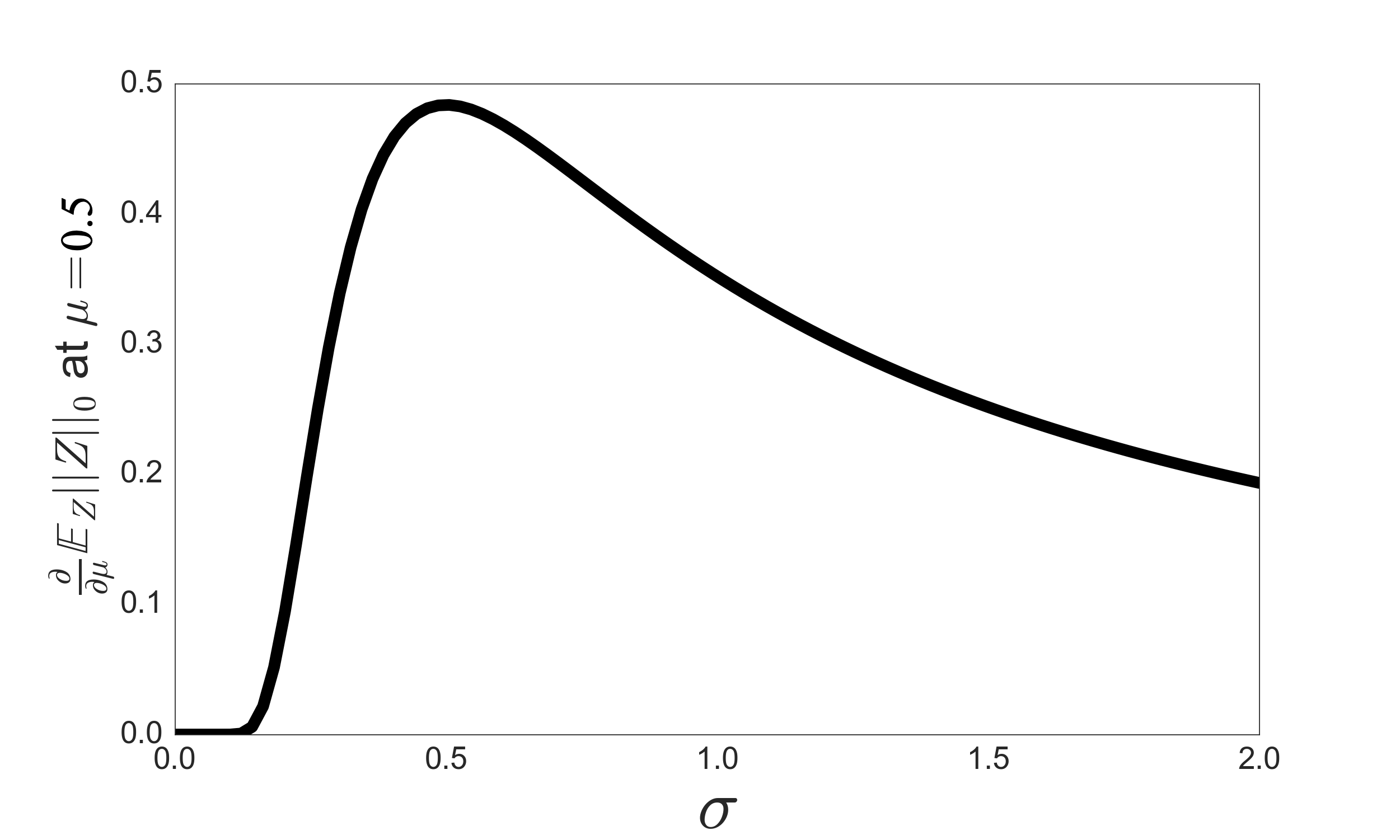}
\end{center}
\caption{The value of $\frac{\partial}{\partial \mu}\mathbb{E}_Z ||\myvec{Z}||_0 \vert _{\mu=0.5} = \frac{1}{\sqrt{2\pi \sigma^2}} e^{-{\frac{1}{8\sigma^2}}}$ for $\sigma = [0.001, 2]$.}
\label{fig:reg_grad}
\end{figure}

To tune $\sigma$, we follow the suggestion in \cite{yamada2020feature}. Specifically, the effect of $\sigma$ can be understood by looking at the value of $\frac{\partial}{\partial \mu_d} \mathbb{E}_Z || \myvec{Z}||_0$. 
In the first training step, $\mu_d$ is $0$. Therefore, at initial training phase, $\frac{\partial}{\partial \mu_d} \mathbb{E}_Z || \myvec{Z}||_0$ is close to $\frac{1}{\sqrt{2\pi \sigma_d^2}} e^{-{\frac{1}{8\sigma_d^2}}}$. To enable sparsification, this term (multiplied by the regularization parameter $\lambda$) has to be greater than the derivative of the loss with respect to $\mu_d$ because otherwise $\mu_d$ is updated in the incorrect direction.
To encourage such behavior, we tune $\sigma$ to the value that maximizes the gradient of the regularization term. As demonstrated in Fig. \ref{fig:reg_grad} this is obtained when $\sigma=0.5$. Therefore, we keep $\sigma=0.5$ throughout our experiments unless specifically noted.
    
\subsection{Linear Regression Example Details}
\label{sec:linear_details}

\subsubsection{Details of the Motivating Example and Extended Evaluations}

First, we describe the data model used in the motivating example (see Section \ref{sec:mot_exp}) and the extended results in section \ref{sec:linear_varying}.

In total, the data matrix $\myvec{X}$ has $2N = 600$ samples where the first $300$ samples (group $1$) are $\text{i.i.d.}$ based on $\mathcal{N}(\myvec{1},0.5\myvec{I})$, and the remaining $300$ $\text{i.i.d.}$ samples (group $2$) are drawn from $\mathcal{N}(-\myvec{1},0.5\myvec{I})$ where $\myvec{I}$ is a $5 \times 5$ identity matrix. $10\%$ of the samples ($60$ data points) are used as validation set, $10\%$ of the samples ($60$ data points) are used as the test set. From the remaining data ($480$ data points), we randomly pick small subsets of samples ($60$,$30$,$18$,$12$,$10$,$6$ samples) as our training sets. We compute $\textnormal{y}$ based on Eq. \ref{eq:linear_regression_syn} in the main text for all the samples.  For the motivating example, the training set has $10$ samples. For the extended evaluations, the training set has $60$,$30$,$18$,$12$,$6$ samples for each case.

For LLSPIN, LASSO, Random Forest, and Neural Network, we optimize each model using $100$ trials of Optuna (a hyper-parameter optimization software \cite{OPTUNA}) on the validation set by minimizing the validation mean squared error, with grids of parameters listed in Table \ref{tab:param_lin}. After Optuna selects the parameters, we test each model's performance on the test set. 

\begin{table}[htb!]
    \centering
    \begin{tabular}{|c|c|}
    \hline
    Parameters & Search Range \\
    \hline
    learning rate (LLSPIN,Neural Net) & [1e-2,2e-1] \\
    \hline
    epochs (LLSPIN,Neural Net) & $\{2000,5000,10000,15000\}$\\
    \hline
    $\lambda$ (LLSPIN) & [1e-3,1e-2] \\
    \hline
    $\alpha$ (LASSO) & [1e-3,5e-1]\\
    \hline
    n\_estimators (RF) & [1,500] \\
    \hline
    max\_depth (RF) & [1,30] \\
    \hline
    min\_samples\_split (RF) & [2,10] \\
    \hline
    \end{tabular}
    \caption{Parameter settings for different models applied to the linear synthetic example}
    \label{tab:param_lin}
\end{table}

For the motivating example where the training size is $10$, we include Localized LASSO and INVASE into comparison. For Localized LASSO, the affinity between samples are computed using the default gaussian kernel from sklearn.metrics.pairwise\\.pairwise\_kernels. We also optimize these $2$ models via 100 trials of Optuna. For Localized LASSO, the grid of $lam\_net$ and $lam\_exc$ is set to be $\{0.001,0.01,0.1,1,5,10,\\50,100\}$, and the grid of number of iterations is $\{100,300,500,1000\}$. For INVASE, the grid of $\lambda$ is [$1e-3$,$1e-2$]. The grid of number of epochs is $\{2000,5000,10000,15000\}$. The grid of learning rate is [$1e-5$,$1e-4$].

In this example, we set the architecture of LLSPIN and the baseline Neural Network to $4$ hidden layers with $100$, $100$, $10$, $1$ neurons in each layer, respectively. For the \textit{gating} network of LLSPIN we use $1$ hidden layer with $10$ neurons. We use full batch training for both LLSPIN and the Neural Network. INVASE is set to have identical architectures.

\subsubsection{Details of the Experiment involving Linear Synthetic Dataset with Unequal Feature Coefficients}

We take all the $480$ remaining data points as training samples for this experiment. $\textnormal{y}$ is computed based on Eq.\ref{eq:linear_regression_syn_uneql} in Section \ref{sec:add_lin} for all the samples. In this example, we set the architecture of LSPIN to $3$ hidden layers with $100$, $10$, $1$ neurons in each layer, respectively. For the \textit{gating} network of LSPIN we use $2$ hidden layer with $100$ neurons in each layer. The activation function of the \textit{prediction} network is set to relu for this particular example. The $\lambda$ is set to $10^{-5}$, the learning rate is set to $0.2$, and the number of epochs is set to $3500$. We use full batch training for this example.

\subsection{Nonlinear Synthetic Datasets Details}
\label{sec:nonl_details}
Here we provide details for reproduction of the example presented in Section \ref{sec:synt}.

\subsubsection{Data Generation and Split}

\textbf{E1}, \textbf{E2}, and \textbf{E3} are adapted from \cite{yoon2018invase} (see Eq. \ref{eq:exp1}, \ref{eq:exp2}, and \ref{eq:exp3}). In each example, we generate the data matrix $\myvec{X}$ with $2000$ samples and $11$ features that are sampled independently from $\mathcal{N}(\myvec{0},\myvec{I})$ where $\myvec{I}$ is an $11 \times 11$ identity matrix. The response is $\textnormal{y} = \myvec{1}_A(\frac{1}{1+\textnormal{Logit}(\myvec{x})} > 0.5)$ where $\myvec{1}_A$ is an indicator function and the $\textnormal{Logit}(\myvec{x})$ for each sample is calculated based on different features depending on the sign of the $11$th feature $x_{11}$. Each Logit is defined based on one of the following equations

\begin{align}
    \textbf{E1:} \qquad \textnormal{Logit} &=
    \begin{cases}
        e^{(\myvec{x}_1 \times \myvec{x}_2)},&\text{if } \myvec{x}_{11}<0,\\ 
        e^{(\sum_{i=3}^{6} \myvec{x}_i^2 - 4)},& \text{otherwise}  \label{eq:exp1}\\
    \end{cases} \\
    \textbf{E2:} \qquad \textnormal{Logit} &=
    \begin{cases}
        e^{(\myvec{x}_1 \times \myvec{x}_2)},&\text{if } \myvec{x}_{11} < 0,\\
        e^{(-10\sin{(0.2\myvec{x}_7)}+|\myvec{x}_8|+\myvec{x}_9+e^{-\myvec{x}_{10}}-2.4)},&\text{otherwise} \label{eq:exp2}\\
    \end{cases} \\
    \textbf{E3:} \qquad \textnormal{Logit} &=
    \begin{cases}
        e^{(\sum_{i=3}^{6} \myvec{x}_i^2 - 4)}, &\text{if } \myvec{x}_{11}<0,\\
        e^{(-10\sin{(0.2\myvec{x}_7)}+|\myvec{x}_8|+\myvec{x}_9+e^{-\myvec{x}_{10}}-2.4)}, &\text{otherwise} \label{eq:exp3}\\
    \end{cases} \\
\end{align}
To evaluate a LSS regime, the number of samples we use is far fewer compared with the number of samples used in experiments conducted in \cite{yoon2018invase}. We split the data and used $90\%$ for training and $10\%$ for testing. $5\%$ of the training set is set aside as a validation set.

Additionally, to further demonstrate LSPIN's robustness on challenging domains, we design a $4^\text{th}$ example (termed \textbf{E4}, see Eq. \ref{eq:exp4}), for which we generate the data matrix $\myvec{X}$ with $2N = 1000$ samples and $4$ features that consists of $2$ sample groups. The first $N$ samples are sampled from $\mathcal{N}(\myvec{1},0.5\myvec{I})$, and the second $N$ samples are sampled from $\mathcal{N}(-\myvec{1},0.5\myvec{I})$ where $\myvec{I}$ is a $4 \times 4$ identity matrix. The response is defined as $\textnormal{y} = \myvec{1}_A(\frac{1}{1 + \textnormal{Logit}(\myvec{x})} >0.5)$, where $\myvec{1}_A$ is an indicator function and $\textnormal{Logit}(\myvec{x})$ depends on features $x_1$ and $x_2$ for the first $N$ samples, and on features $x_3$ and $x_4$ for the remaining $N$ samples. To make the classification task harder, we add other $46$ nuisance features (irrelevant for the prediction task) sampled from $\mathcal{N}(\myvec{0},0.5\myvec{I})$ where $\myvec{I}$ is an $46 \times 46$ identity matrix.

\begin{align}
     \textbf{E4:} \qquad \textnormal{Logit} &=
        \begin{cases}
            e^{(\myvec{x}_1\times\myvec{x}_2 - 0.9)},&\text{if first N samples}\\
            e^{(\myvec{x}_3^2 + \myvec{x}_4^2 - 2.5)},&\text{otherwise} \label{eq:exp4} \\
    \end{cases}
\end{align}

For \textbf{E4}, we split the data and use $95\%$ for training and $5\%$ for testing. $10\%$ of the training set is set aside as validation set.

Lastly, to evaluate our model in the nonlinear regression regime, we design a moving XOR dataset as the $5$th example (\textbf{E5},see Eq. \ref{eq:exp5}). Specifically, we generate the data matrix $\myvec{X}$ with $3N=2100$ samples and $20$ features, where each entry is sampled from a fair Bernoulli distribution ($P(x_{ij} = 1) = P(x_{ij} = -1) = 0.5$). Then we add an additional feature $x_{21}$ for each sample where $x_{21} = -1$ for the first $N$ samples, $x_{21} = 0$ for the second $N$ samples, $x_{21} = 1$ for the last $N$ samples. Based on the value of $x_{21}$, the response variable $\textnormal{y}$ for different samples will have different subset of features, as defined in  Eq. \ref{eq:exp5}.

\begin{align}
    \textbf{E5:} \qquad \textnormal{y} &=
    \begin{cases}
        \myvec{x}_1 \times \myvec{x}_2 + 2\myvec{x}_{21},& \text{if } \myvec{x}_{21} = -1,\\ 
       \myvec{x}_2 \times \myvec{x}_3 + 2\myvec{x}_{21},& \text{if } \myvec{x}_{21} = 0,\\
        \myvec{x}_3 \times \myvec{x}_4 + 2\myvec{x}_{21},& \text{if } \myvec{x}_{21} = 1,\\
    \end{cases}  \label{eq:exp5}
\end{align}

For \textbf{E5}, the training set has $1500$ samples, the validation and test have $300$ samples each.

\subsubsection{Training Procedures and Hyper-parameter Tuning and Settings}

For these $5$ experiments, we optimize each model on the validation set (minimizing classification error for classification and mean squared error for regression) using Optuna and evaluate the optimized models on the test sets.
For the neural network based methods, the F1-score of the selected features is also evaluated on the test set.

For lasso, we optimize the $l_1$ regularization parameter with $20$ trials and the grid range is $[1e-2,1e3]$. For Random Forest, we optimize the number of estimators, max\_depth, and min\_samples\_split with $100$ trials and the corresponding grid ranges are $[1,500]$, $[1,30]$, $[2,10]$.

For LLSPIN/LSPIN/INVASE/L2X/REAL-x, the parameter settings and grids are listed in Table \ref{tab:nonl_parameters}. The number of hidden layers and nodes are identical for these models. For \textbf{E2}, \textbf{E3}, \textbf{E4}, we use $2$ hidden layers with $200$ nodes each for the prediction network architecture. We use $2$ layers with $100$ nodes each for the gating network of LLSPIN/LSPIN and the selector network of INVASE/L2X/REAL-x. For \textbf{E1}, we add one layer with $200$ nodes to the prediction architecture for all models and one layer with $100$ nodes to the gating network of LLSPIN/LSPIN and selector network of INVASE/L2X/REAL-x. For \textbf{E5}, we use $3$ hidden layers with $500$, $100$, $1$ nodes each for the prediction network architecture. We use $1$ layer with $100$ nodes for the gating network of LLSPIN/LSPIN and the selector network of INVASE/L2X. For \textbf{E5}, we use leaky relu as the activation function for the prediction network of LSPIN.

To visualize the selected features for LSPIN (Fig. \ref{fig:nonl_combo}), we repeated the optimization procedure $5$ times and plotted the average gate values on the test set.

\begin{table}[htb!]
    \centering
    \begin{adjustbox}{max width=1 \textwidth,min height = 0.6 in, valign=l}
    \begin{tabular}{|c|c|c|c|c|}
    \hline
           & Batch Size & Number of Epochs & Learning Rate & $\lambda$/$k$ \\
    \hline
    \textbf{E1} & Full & $10000$ (LLSPIN/LSPIN/INVASE) & 1e-1 (LLSPIN/LSPIN) & [2e-1,3e-1]/  \\

    & & 
    \{$1000$,$3000$,$5000$,$7000$,$9000$\} (L2X)   & 1e-4 (INVASE) &  \{1,2,3,4,5,6,7\} \\

    & & \{$500$,$1000$,$2000$,$5000$,$10000$\} (REAL-x)  &  [1e-5,1e-2] (L2X, REAL-x) & \\
    
    \hline
    
    \textbf{E2} & Full & $10000$ (LLSPIN/LSPIN/INVASE) 
   & 1e-1 (LLSPIN/LSPIN) & [0.1,0.15]/\\

    & &  \{$1000$,$3000$,$5000$,$7000$,$9000$\} (L2X)  & 1e-4 (INVASE)& \{1,2,3,4,5,6,7\}\\

    &  & \{$500$,$1000$,$2000$,$5000$,$10000$\} (REAL-x)  & [1e-5,1e-2] (L2X, REAL-x) & \\
    \hline

    \textbf{E3} & Full & $3000$ (LLSPIN/LSPIN/INVASE) 
     &  1e-1 (LLSPIN/LSPIN) & [0.15,0.2]/\\

     &  & 
     \{$1000$,$3000$,$5000$,$7000$,$9000$\} (L2X)   &   1e-4 (INVASE)  & 
     \{1,2,3,4,5,6,7\} \\

    & & \{$500$,$1000$,$2000$,$5000$,$10000$\} (REAL-x) &  [1e-5,1e-2] (L2X, REAL-x) &  \\
    \hline

    \textbf{E4} & Full & $\{1000,1200\}$ (LLSPIN/LSPIN/INVASE)   
    & [3e-2,5e-2] (LLSPIN/LSPIN) & [1.33,1.35]/\\
    
    &  &    
    \{$1000$,$3000$,$5000$,$7000$,$9000$\} (L2X)  & [1e-5,1e-4] (INVASE) & \{1,2,3,4,5,6,7\}\\
    
    & &  \{$500$,$1000$,$2000$,$5000$,$10000$\} (REAL-x) & [1e-5,1e-2] (L2X, REAL-x) & \\
    \hline
    \textbf{E5} & Full & \{$2000$,$3000$,$5000$,$7000$\}  & [1e-2,1e-1] (LLSPIN/LSPIN) & $1$/\\
    
    & & (LLSPIN/LSPIN/INVASE) & [1e-5,1e-3] (INVASE) & \{1,2,3,4,5,6,7\}\\
    
    & & \{$1000$,$3000$,$5000$,$7000$,$9000$\} (L2X) & [1e-5,1e-2] (L2X) & \\
    
    \hline
    \end{tabular}
    \end{adjustbox}
    \caption{Parameter settings and grids for the $5$ nonlinear experiments. For LLSPIN/LSPIN/INVASE, we run $5$ Optuna trials for \textbf{E1}, \textbf{E2}, \textbf{E3} and $20$ trials for \textbf{E4} and \textbf{E5}. For L2X and REAL-x, we run $100$ trials across datasets. Note that LLSPIN/LSPIN use SGD Optimizer and others use Adam optimizer.}
    \label{tab:nonl_parameters}
\end{table}

To evaluate a fair comparison of the prediction performance, we sparsify the input to the prediction network of REAL-x by multiplying the original information with the output from the selector network.

For TabNet, the grid of the regularization parameter $\lambda_{sparse}$ is set to \{$0.0001$, $0.001$,$0.01$,$0.1$,$0.2$,$0.3$,$0.5$\}. The grids of $n_d$ and $n_a$ are both \{$8$,$16$,$24$,$32$,$64$,$128$\}. The grid of number of steps is \{$3$,$4$,$5$,$6$,$7$,$8$,$9$,$10$\}. The grid of $\gamma$ is \{$1.0$,$1.2$,$1.5$,$2.0$\}. The grid of learning rate is \{$0.005$,$0.01$,$0.02$,$0.025$\}, and the grid of momentum is \{$0.6$,$0.7$,$0.8$,$0.9$,$0.95$,$0.98$\}. The scheduler function is set to StepLR and the grids of the corresponding step\_size and decay rate are \{$500$,$2000$,$8000$,$10000$,$20000$\} and \{$0.4$,$0.8$,$0.9$,$0.95$\}, respectively. The grid of max number of epochs is \{$4000$,$10000$,$20000$\}. The batch size is set to full batch and the virtual batch size is set to $5$. The early stopping patience is set to $30$ epochs. Other parameters are set to be default. We optimize the model on each dataset with $100$ trials of Optuna.

\subsection{Real-world Datasets Details}
\label{sec:real_world_details}
Here we provide details for reproduction of the example presented in Section \ref{sec:real_world}.

\subsubsection{PBMC Dataset Preprocessing Steps} 
The purified Peripheral Blood Mononuclear Cells (PBMC) dataset is collected from \cite{pbmcP}, in which the data matrix has cells as samples, genes as features, and each entry represents the number of mRNAs expressed from the corresponding gene of that cell. This raw data matrix is first filtered (cells that have less than 400 expressed genes are excluded, and genes that are expressed in less than 100 cells are excluded) and normalized by the library size (total number of mRNAs expressed per cell). We then exclude the non-protein-coding genes and retain only cells that belong to the following $4$ cell types: memory T cells, naive T cells, regulatory T cells, naive cytotoxic T cells.

We use $34,115$ cells ($90\%$ of the data) to select the $2000$ most variable genes and use the remaining $3,791$ cells ($10\%$ of the data) with these $2000$ genes as the final processed dataset. Then, we split training/test/validation sets as described in the following subsection.

\subsubsection{Training Procedures} 

In this section, we introduce our training procedures for the real-world  LSS datasets. Specifically, for the BASEHOCK, RELATHE, PCMAC, and PBMC datasets, $5\%$ of each dataset is set aside as a validation set. Let us denote the remaining $95\%$ of the data as $\myvec{\Bar{X}}$. We split $\myvec{\Bar{X}}$ into $5$ non-overlapping folds. 

We train each model on $1$ fold of $\myvec{\Bar{X}}$ and test it on the remaining non-overlapping $4$ folds of $\myvec{\Bar{X}}$. The hyper-parameters are optimized on the validation set via Optuna (50 trials for the neural network-based methods and tree-based methods and five trials for INVASE due to long computation time) based on the model trained on a single fold and tested on the remaining $4$ non-overlapping folds. These (fixed) hyper-parameters are then used to train the model on the second fold and test it on the remaining $4$ non-overlapping folds. Similarly, we use these fixed hyper-parameters to train models for folds No.3, No.4, and No.5, and each time test these models on the remaining $4$ non-overlapping folds. This training and testing procedure is repeated for several regularization parameters; then, we report the best average performance for each method.  

Since COLON and TOX-171 are of extreme LSS, we use a grid of regularization parameters for each method and identify the best average performance (test accuracy and the number of selected features) across ten runs (using 80\% of the samples for training and 20\% for testing).

The regularization parameters are tuned to select fewer than $50$ features, except for XGBoost when we applied it to the PBMC dataset. In this case, the minimum possible number of features chosen by XGBoost was $64$. For local methods including LLSPIN, LSPIN, and INVASE, L2X, TabNet, and REAL-x, the average (over the five folds) median (over the training samples) number of selected features is reported. 

\subsubsection{Hyper-parameter Tuning and Settings}
For LLSPIN/LSPIN/STG/Neural Network model, the prediction network architecture is set to 3 hidden layers with 100, 50, 30  neurons, respectively, for all the datasets. The gating network for LLSPIN/LSPIN models is set to one single layer with 500 neurons for BASEHOCK, RELATHE, PCMAC, PBMC datasets, and two hidden layers with 100 and 10 neurons respectively for the COLON and TOX-171 datasets. For INVASE/L2X/REAL-x, the network architecture is set to default. The predictor network has two hidden layers and 200 neurons on each layer, and the selector network has two hidden layers and 100 neurons on each layer. 

For these neural network-based methods, the grids of regularization parameter $\lambda$ are listed in Table \ref{tab:real_data_grids}, along with the grids of learning rate and epochs that are optimized via Optuna for the BASEHOCK, RELATHE, PCMAC, PBMC datasets, and the settings of learning rate and epochs in the COLON and TOX-171 datasets. We use full batch training for all neural network-based methods for all $6$ datasets, except for TabNet, where we set batch size and virtual batch size to be $100$ and $10$ for BASEHOCK/RELATHE/PCMAC/PBMC and $20$ and $4$ for COLON and TOX-171.

\begin{table}[htb!]
    \centering
    \begin{adjustbox}{max width=1 \textwidth,min height = 0.5 in, valign=l}
    \begin{tabular}{|c|c|c|c|c|}
 
    \hline
    Dataset & Method & $\lambda$/$k$/$\lambda_{sparse}$ & Learning Rate & Number of Epochs   \\
    \hline
     \multirow{6}{*}{\shortstack{BASEHOCK\\RELATHE\\PCMAC\\PBMC}} & LLSPIN & [1,10] & [1e-2,1e-1] & $\{1000,3000,5000,7000,9000\}$ \\\cline{2-5}
                          & LSPIN & [1,10] & [1e-2,1e-1] & $\{1000,3000,5000,7000,9000\}$  \\\cline{2-5}
                          & STG (l) & [1,10] & [1e-1,2e-1]  & $\{3000,5000,7000,9000\}$  \\\cline{2-5}
                          & STG (n) & [1,10] & [1e-1,2e-1]  & $\{3000,5000,7000,9000\}$ \\\cline{2-5}
                          & Neural Net  & None & [1e-2,1e-1]  & $\{1000,3000,5000,7000,9000\}$ \\\cline{2-5}
                          & INVASE & $\{1,5,10\}$ & [1e-5,1e-4] & $10000$ \\\cline{2-5}
                          & L2X & $\{1,5,10\}$ & [1e-5,1e-2] & $\{1000,3000,5000,7000,9000\}$\\\cline{2-5}
                          & TabNet & $\{0.0001,0.001,0.01,0.1,0.2,0.3,0.5\}$ & [0.005,0.025] & $\{2000,4000,6000,8000,10000\}$\\\cline{2-5}
                          & REAL-x & $\{10,30,50,70,90\}$ & [1e-5,1e-2] & $\{100,200,500,1000,2000\}$\\\hline
     \multirow{6}{*}{\shortstack{COLON}} & LLSPIN & [1,2] & 0.1 & 7000  \\\cline{2-5}
                            &LSPIN & [1,2] & 0.05 & 7000  \\\cline{2-5}
                            & STG (l)     & [1,2] & 0.5 & 7000  \\\cline{2-5}
                            & STG (n)     & [1,2] & 0.5 & 7000  \\\cline{2-5}
                            & Neural Net  & None & 0.1 & 7000  \\\cline{2-5}
                            & INVASE      & $\{1,1.5,2\}$ & 0.0001 & 10000  \\\cline{2-5}
                            & L2X         & $\{1,3,5,7,9,10\}$ & 0.0001 & 10000 \\\cline{2-5}
                            & TabNet      &  $\{0.0001,0.001,0.01,0.1,0.2,0.3,0.5\}$ & 0.0001 & 10000\\\cline{2-5}
                            & REAL-x      & $\{1,5,10,30,50\}$  & 0.0001 & 1000\\\hline
     \multirow{6}{*}{\shortstack{TOX-171}} & LLSPIN & [1,2] & 0.1 & 7000  \\\cline{2-5}
                            &LSPIN & [1,2] & 0.05 & 7000  \\\cline{2-5}
                            & STG (l)     & [1,10] & 0.5 & 7000  \\\cline{2-5}
                            & STG (n)     & [1,10] & 0.5 & 7000  \\\cline{2-5}
                            & Neural Net  & None & 0.1 & 7000  \\\cline{2-5}
                            & INVASE      & $\{1,1.5,2\}$ & 0.0001 & 10000  \\\cline{2-5}
                            & L2X         & $\{1,3,5,7,9,10\}$ & 0.0001 & 10000 \\\cline{2-5}
                            & TabNet      &  $\{0.0001,0.001,0.01,0.1,0.2,0.3,0.5\}$ & 0.0001 & 10000\\\cline{2-5}
                            & REAL-x      & $\{1,5,10,30,50\}$  & 0.0001 & 1000\\\hline
   \end{tabular}
    \end{adjustbox}
    \caption{Parameter settings for the neural network based methods on the real-world data. Note that INVASE/L2X/TabNet/REAL-x use Adam Optimizer and others use SGD optimizer. For the regularization parameter $\lambda$, the grid size for LSPIN and STG models is $5$ on BASEHOCK/RELATHE/PCMAC/PBMC datasets and $20$ on COLON and TOX-171 datasets.
  }
    \label{tab:real_data_grids}
\end{table}

We studied the TOX-171 dataset, setting the grid of $\lambda$ of the STG models in the range of  [1,2]. We observed that in this range, the number of features was too high; therefore we extended the range to [1,10] for the STG models.

For LASSO and SVC, the grid for their regularization parameter $c$ is set to [1e-3,1e-1] for the BASEHOCK, RELATHE, PCMAC, PBMC datasets, and [1e-2,1e3] for the COLON and TOX-171 datasets.


For Random Forest and XGBoost, we use number\_of\_estimators to replace the regularization parameter proposed in the previously training procedures. The grid of number\_of\_estimators is  \{1,5,10,20,30,50,100,200,500,1000\} for both methods when we applied them to the  BASEHOCK, RELATHE, PCMAC, PBMC datasets, and is \{1,2,3,4,5,8,10,15,20,25,30,40,50,60,70,80,100,200,500,1000\} when we applied them to the COLON and TOX-171 datasets. 


Other parameter settings for XGBoost are as follows: For the  BASEHOCK, RELATHE, PCMAC, PBMC datasets, we optimize max\_depth  via Optuna with grid range [1,10]. For the COLON and TOX-171 datasets, we set max\_depth to $10$.

Other parameter settings for Random Forest are as follows: For the  BASEHOCK, RELATHE, PCMAC, PBMC datasets, we optimize max\_depth and \\minimum\_samples\_to\_split via Optuna with grid range [1,10] and [2,50], respectively. For the COLON and TOX-171 datasets, we set max\_depth to $10$ and minimum\_samples\_to\_split to $5$.

To evaluate a fair comparison of the prediction performance, we sparsify the input to the prediction network of REAL-x by multiplying the original information with the output from the selector network.

For TabNet, the early stopping patience is set to $30$ epochs. Other parameters are set to be Default.

In the MNIST experiment we use a batch size of $100$ with a learning rate of $0.1$ and train for $300$ epochs. When $\lambda_2>0$, we use a warm up procedure where we first train the model for $200$ epochs with $\lambda_2=0$ and then increase $\lambda_2$. We observe that this stabilizes the training procedure.
\subsection{Cox Proportional Hazard Models for Survival Analysis Details} 

Here we provide details for reproduction of the example presented in Section \ref{sec:cox}.

\subsubsection{SEER Dataset Preprocessing Steps}

The data for this study were collected from the Surveillance, Epidemiology, and End Results (SEER) public datasets \cite{national2011surveillance}. Female patients, ages 25-85, diagnosed with histologically confirmed non-metastatic breast cancer between Jan 1, 2000, and Dec 31, 2016, are included. Patients with metastatic disease and those with missing data on stage, T grade, number of positive nodes, number of T nodes, vital status, and survival time are excluded. Only those patients who underwent surgery and had a known tumor size of less than 200 mm are included. Patients with bilateral breast cancer, inflammatory disease, and in-situ tumor are excluded. We use one-hot encoding for the categorical variables and drop features with unknown/unspecified values. Continuous variables are z-scored. We further add $3$ random uniform variables as noise to the data. In total, we have $538,315$ patients and $55$ features after processing.
\subsubsection{Training Procedures and Hyper-parameter Tuning and Settings}

We apply a training procedure similar to the one we used to the BASEHOCK, RELATHE, PCMAC, PBMC datasets for the SEER data.

$5\%$ of each dataset is set aside as a validation set. Let us denote the remaining $95\%$ of the data as $\myvec{\Bar{X}}$. We split $\myvec{\Bar{X}}$ into $10$ non-overlapping folds.

We train each model on $1$ fold of $\myvec{\Bar{X}}$ and test it on the remaining non-overlapping $9$ folds of $\myvec{\Bar{X}}$. The hyper-parameters are optimized on the validation set via Optuna
($50$ Optuna trials for the neural network methods on the learning rate and epochs as in Table \ref{tab:param_surv}) based on the model trained on a single fold and tested on the remaining $9$ non-overlapping folds. These (fixed) hyper-parameters are then used in training the model on the second fold and testing it on the remaining $9$ non-overlapping folds. Similarly, we use these fixed hyper-parameters to train models for folds No.3, No.4, No.5, ..., No.10, and each time test these models on the remaining $9$ non-overlapping folds. We then compute the average (over the ten folds) performance (test concordance index and the number of selected features). This training and testing procedure is repeated for several regularization parameters to produce the results in the interpolation plot Fig.\ref{fig:cox_result} (Left). We note that Random Survival Forest selects almost all the features over different n\_estimators parameters as shown as an isolated interpolation point in Fig.\ref{fig:cox_result} (Left).

\begin{table}[htb!]
    \centering
    \begin{tabular}{|c|c|}
    \hline
    Parameters & Range \\
    \hline
    learning rate (all Neural Network methods) & [1e-2,1] \\
    \hline
    epochs (all Neural Network methods) & $\{500,1000,2000\}$\\
    \hline
    $\lambda$ (COX-LLSPIN, COX-LSPIN) & [1e-9,1e-5] \\
    \hline
    $\lambda$ (COX-STG(Linear/Nonlinear)) & [1e-3,1e-1]\\
    \hline
    $\alpha$ (COX-LASSO) & [1e-7,1] \\
    \hline
    n\_estimators (Random Survival Forest) & \{1,10,100,500,1000\} \\
    \hline
    \end{tabular}
    \caption{Parameter settings for different models for the survival analysis example. For COX-LLSPIN/COX-LSPIN/COX-STG, the regularization parameter is $\lambda$. For COX-LASSO, the regularization parameter is $\alpha$. For Random Survival Forest, we use n\_estimators to replace the regularization parameter.}
    \label{tab:param_surv}
\end{table}

In this example, we set the nonlinear neural network methods (COX-LSPIN, COX-STG(Nonlinear), DeepSurv) to 3 hidden layers with 100, 30, and 5 neurons, respectively. The linear neural network methods (COX-LLSPIN, COX-STG(Linear)) have no hidden layers. The gating network of LLSPIN/LSPIN is set to 1 hidden layer with 300 neurons. For the local methods, including COX-LLSPIN and COX-LSPIN, the average (over the $10$ folds) median (over the training samples) number of selected features is reported.  

\subsection{Single Nucleus RNA-seq Dataset Details}

\label{sec:snrna_details}

\subsubsection{Data Preprocessing and Split}

Similar to the scRNA-seq data,  in the data matrix, the samples are cells, the features are the genes, and each entry represents the number of mRNAs expressed from the corresponding gene of that cell. The cells are first filtered based on the number of genes that have non-zeros values (lower threshold is $500$ and upper threshold is $7500$) and then filtered based on the mitochondrial ratio ($10$\%). The data are imputed using ALRA \cite{linderman2022zero}. 

We randomly sample $1000$ cells of each type (Microglia cells and Oligodendrocyte Precursor Cells (OPC)), then use $50\%$ of the data to 
select $100$ most variable genes that are not correlated with \textit{ITGAM} and \textit{PDGFRA} as noisy genes. For the remaining data, we keep these $100$ genes along with \textit{ITGAM} and \textit{PDGFRA} as features and use $80\%/10\%/10\%$ as the train/validation/test split. 

\subsubsection{Hyper-parameter Tuning and Settings}

For each model in Table \ref{tab:sc_f1}, we optimize over a grid of the corresponding regularization parameter by minimizing the classification error (if two models have the same performance, the sparse one will be chosen). For LLSPIN/INVASE/REAL-x, the grid of $\lambda$ is $\{1,10,20,30,40,50\}$. For L2X, the grid of $k$ is $\{1,2,3,4,5,6,7\}$. For TabNet, the grid of $\lambda_{sparse}$ is $\{0.5,1,5,10,20,50\}$.

For LLSPIN, the prediction network is $2$ layers with $200$ and $100$ nodes each. The gating network is a $1$ layer with $100$ nodes. The learning rate is set to be $0.1$. The number of epochs is $2000$. The batch size is set to full batch training. The standard deviation ($\sigma$) of the Gaussian reparameterization is set to $1$.

For INVASE/L2X/REAL-x, the network architecture is set to default. The predictor network has two hidden layers and 200 neurons on each layer, and the selector network has two hidden layers and 100 neurons on each layer. The learning rate is set to be $1e-3$, the number of epochs is $2000$.

For TabNet, the learning rate is set to $1e-3$, and the number of epochs is $2000$. The batch size is $100$, and the virtual batch size is $10$. The early stopping patience is $30$ epochs. Other parameters are set as default.

\section{Strengths and Limitations}
\label{sec:limit}
The proposed model leads to an interpretable prediction model that can handle datasets of low sample size (LSS). Our results demonstrate that local sparsity tied with a linear model can be a robust classifier on real biological datasets. As for the societal impact, we don't know the effect of adversarial examples on the hazards model.

Currently, the sparsity of our model is tuned via a regularization parameter $\lambda_1$ while stability is tuned via $\lambda_2$ (see Eq. \ref{eq:reg}). In certain setting, tuning these parameters could be a demanding task, since it involves balancing with the main loss term. In the future, we plan to explore a more flexible mechanism for local feature selection. One possible way that this could be realized is using a concrete layer as proposed in \cite{ConcreteAuto}.

\end{document}